\documentclass{article}

\usepackage{color,xcolor}
\usepackage{epsfig}
\usepackage{graphicx}

\usepackage{adjustbox}
\usepackage{array}
\usepackage{booktabs}
\usepackage{colortbl}
\usepackage{float,wrapfig}
\usepackage{hhline}
\usepackage{multirow}
\usepackage{subcaption} % issues a warning with CVPR/ICCV format

\usepackage{amsmath,amsfonts,amsthm,amssymb}
\usepackage{bm}
\usepackage{nicefrac}
\usepackage{microtype}

\usepackage{changepage}
\usepackage{extramarks}
\usepackage{fancyhdr}
\usepackage{lastpage}
\usepackage{setspace}
\usepackage{soul}
\usepackage{xspace}

\usepackage[pagebackref=true,breaklinks=true,colorlinks,citecolor=gray]{hyperref}
\usepackage{url}

\usepackage{algorithm, algorithmic}
\usepackage{enumerate}
\usepackage{todonotes} % conflict with CVPR/ICCV format

\usepackage{titlesec}
\usepackage{makecell}

\usepackage{amsmath,amsthm,amssymb}
\usepackage{mathtools}
\usepackage{paralist}
\newcolumntype{L}[1]{>{\raggedright\let\newline\\\arraybackslash\hspace{0pt}}m{#1}}
\newcolumntype{C}[1]{>{\centering\let\newline\\\arraybackslash\hspace{0pt}}m{#1}}
\newcolumntype{R}[1]{>{\raggedleft\let\newline\\\arraybackslash\hspace{0pt}}m{#1}}

\newcommand{\fig}[1]{Figure~\ref{#1}}

\newcommand{\ignore}[1]{}
\newcommand{\norm}[1]{\lVert#1\rVert}

\makeatletter
\DeclareRobustCommand\onedot{\futurelet\@let@token\@onedot}
\def\@onedot{\ifx\@let@token.\else.\null\fi\xspace}

\makeatother

\definecolor{MyDarkBlue}{rgb}{0,0.08,1}
\definecolor{MyDarkGreen}{rgb}{0.02,0.6,0.02}
\definecolor{MyDarkRed}{rgb}{0.8,0.02,0.02}
\definecolor{MyDarkOrange}{rgb}{0.40,0.2,0.02}
\definecolor{MyPurple}{RGB}{111,0,255}
\definecolor{MyRed}{rgb}{1.0,0.0,0.0}
\definecolor{MyGold}{rgb}{0.75,0.6,0.12}
\definecolor{MyDarkgray}{rgb}{0.66, 0.66, 0.66}

\newcommand{\myitem}{\vspace{-6pt}\item}
\titlespacing*{\section}{0pt}{0pt plus 2pt minus 2pt}{0pt plus 2pt minus 2pt}
\titlespacing\subsection{0pt}{0pt plus 1pt minus 1pt}{0pt plus 1pt minus 1pt}
\DeclarePairedDelimiterX{\infdivx}[2]{(}{)}{%
  #1\;\delimsize|\delimsize|\;#2%
}

\DeclareMathOperator{\E}{\mathbb{E}}

\usepackage{comment}

\usepackage{hyperref}

\usepackage[accepted]{icmlw2019generalization}
\icmltitlerunning{An Empirical Study on Hyperparameters and their Interdependence for RL Generalization}
\hypersetup{draft}
\begin{document}
\twocolumn[
\icmltitle{An Empirical Study on Hyperparameters and their Interdependence for RL Generalization}



\icmlsetsymbol{equal}{*}

\begin{icmlauthorlist}
\icmlauthor{Xingyou Song}{goog}
\icmlauthor{Yilun Du}{mit}
\icmlauthor{Jacob Jackson}{waterloo}
\end{icmlauthorlist}

\icmlaffiliation{waterloo}{University of Waterloo}
%\icmlaffiliation{openai}{OpenAI, San Francisco, California, USA}
\icmlaffiliation{goog}{Google Brain}
\icmlaffiliation{mit}{MIT}
%\icmlaffiliation{berkeley}{Computer Science Department, UC Berkeley, California, USA}
\icmlcorrespondingauthor{Xingyou Song}{xingyousong@google.com}

\icmlkeywords{reinforce, theory, learning, gradient, cosine, generalization, batchnorm, coinrun, LQR, observation, Machine Learning, ICML}

\vskip 0.3in
]



\printAffiliationsAndNotice{Work done while all authors were at OpenAI.}  % leave blank if no need to mention equal contribution

\begin{abstract}
Recent results in Reinforcement Learning (RL) have shown that agents with limited training environments are susceptible to a large amount of overfitting across many domains. A key challenge for RL generalization is to quantitatively explain the effects of changing parameters on testing performance. Such parameters include architecture, regularization, and RL-dependent variables such as discount factor and action stochasticity. We provide empirical results that show complex and interdependent relationships between hyperparameters and generalization. We further show that several empirical metrics such as gradient cosine similarity and trajectory-dependent metrics serve to provide intuition  towards these results.
\end{abstract}

\section{Introduction}
\label{introduction}

Although reinforcement learning (RL) has allowed automatic solving of a variety of environments with rewards, the topic of generalization in deep reinforcement learning (RL) has recently been very prominent, due to the brittleness witnessed in policies in a variety of environments. 

One framework used to study RL generalization is to treat it analogous to a classical supervised learning problem - i.e. train jointly on a finite "training set", and check performance on the "test set" as an approximation to the population distribution. Normally, statistical learning theory only provides measurements of generalization performance \textit{at the end}, based on a probabilistic bounding approach using the complexity of the classifier.

This however, ignores many trajectory dependent factors. Many different RL-specific parameters may affect RL performance \textit{during training}, including $\gamma$ - the discount factor, action stochasticity, and other network modifications such as Batch Normalization. \textbf{Trajectory Dependent} methods of analysis thus are still important (which include direction of gradient, Gradient Lipschitz smoothness, and optimization landscapes), and seek to understand training behavior during gradient descent. 

However, one large caveat in RL is simply the inherent noisiness in evaluating the objective function, which can easily make landscape visualization such as \cite{visualizingloss, madrypolicygradient}, intractible and expensive for larger scale datasets, such as CoinRun \cite{coinrun}. Instead, we must infer properties about the loss landscape and training from observing summary metrics. The main metric we will use is the gradient cosine similarity between training and testing sets during training. 

While there has been work on understanding what may happen to RL generalization when \textit{one} hyperparameter is changed \cite{dependencehorizon, entropy_regularization}, the theory of RL hyperparameters's coupling effects on generalization is not fully understood. One question raised is if these factors are \textit{independent to each other} - for instance, if we add entropy bonuses to improve generalization, should we ignore tuning other hyperparameters like $\gamma$? Or for instance, if the MDP family is more noisy and stochastic, should we ignore tuning parameters such as the mini-batchsize? 

We provide separate experimental results measuring parts of the policy gradient optimization process, to show that 

\begin{itemize}
    \myitem Many hyperparameters are \textit{not} orthogonal to each other with respect to generalization performance. In fact, the addition of one hyperparameter may completely change the monotonicity of another hyperparameter with respect to generalization.
    \myitem Different regularizations do not affect the training process the same way, and gradient cosine similarity is one main metric to show their effects. 
\end{itemize}

\section{Notation and Methods}
To formalize our supervised learning analogy to the RL setup, let $\Theta$ be a distribution over parameters $\theta$ that parametrize an MDP family $\{\mathcal{M}_\theta : \theta \in \Theta\}$. Each $\theta$ parametrizes some state space, action space, reward, transitions, and observation function, with $\mathcal{M}_\theta = (\mathcal{S}_\theta, \mathcal{A}_\theta, r_\theta, \mathcal{T}_\theta, S_{0,\theta}, w_{\theta})$. An appropriate train and test set can then be created by randomly sampling $\theta \sim \Theta$ and training or evaluating on $\mathcal{M}_\theta$.

For sake of notational simplicity, we denote $s$ as the observation rather than $w_{\theta}$. The standard policy gradient \cite{reinforce} without a discount factor where $\tau = (s_{1},a_{1},...,s_{T}, a_{T})$ is the gradient with respect to the true objective:
\begin{equation}\nabla_{\phi} R(\pi_{\phi}) = \E_{\tau} \left[ \left( \sum_{t=1}^{T} r_{t} \right) \left(\sum_{t=1}^{T} \nabla_{\phi}  \log  \pi_{\phi}(a_{t} | s_{t}) \right) \right] \end{equation}
More recent RL algorithm such as PPO \cite{ppo} optimize a surrogate objective $R^{PPO}(\beta) = L^{CLIP}(\phi) - c_{1} L^{VF}(\phi) + c_{2} S(\pi_{\phi})$, where $L^{CLIP}$ is the clipped advantage ratio, $L^{VF}$ is the value function error, and $S$ is entropy.

These surrogate losses inherently affect our definition of the "gradient cosine similarity" (GCS) between training and test gradients and what the GCS measures. With basic policy gradient this is simply the normalized dot product between $\nabla_{\phi} R_{train}(\pi_{\phi}) $ and $\nabla_{\phi}R_{test}(\pi_{\phi})$ which is unbiased, but with algorithms such as PPO, the GCS becomes biased. On datasets that practically require PPO, we minimize its subtle effects on the GCS by using the same algorithm hyper-parameters for training and testing.

We briefly explain some metrics and what they measure: 
\begin{itemize}
\item GCS: If training gradient is aligned more with true gradient, larger learning rates may be allowed. Furthermore,  \cite{batchnorm_madry} uses GCS as an approximation of the Hessian and smoothness of training landscape - having this oscillate too highly suggests a sharp minimizer, which is bad for generalization. 

\item Variance, Gradient Norm, and $\ell_{2}$ weight norm: The variance of the gradient estimation is dependent on both the batch-size as well as the Lipschitz smoothness of the policy gradient with respect to policy parameters, i.e. $\norm{\nabla_{\phi}R_{S}(\pi_{\phi}) - \nabla_{\phi}R_{S}(\pi_{\phi'})} \le L \cdot \norm{\phi - \phi'}$. Gradient Norm is a practical estimate of the Lipschitz constant between different total rewards, and $\ell_{2}$ is a rough estimate of the complexity of the policy.
\end{itemize}

\section{MDP's for Experimentation}
\cite{deeprl_matters,madrypolicygradient} establish deep RL experimentation is inherently noisy, and joint training only increases the variance of the policy gradient is increased further from sampling from different environments. The raw versions of ALE and Mujoco environments do not strongly follow distributional sampling of levels. We instead use both synthetic (RNN-MDP) and real (CoinRun) datasets, with policy gradient PPO as the default algorithm due to its reliability (hyperparameters in Appendix, \ref{hyper}).

\subsection{RNN-MDP}
To simulate non-linear dynamics for generalization, we fix an RNN to simulate $\mathcal{T}$. The RNN's (input, hidden state, output) correspond to the (action of a policy, underlying MDP state,  observations $w$) with scalar rewards obtained through a nonlinear map of $w$. The initial state will start from $s_{0}(\theta)$. Thus the the parameter $\theta$ will only be a seed to generate initial state. This setting is done in order to use a constant transition function, to guarantee that there always exists an optimal action for any state regardless of $\theta$. By allowing control over state transitions, we may study the effects of stochasticity in the environment, by adding Gaussian noise into the hidden state at each time, i.e. $s_{t}^{actual} = s_{t}^{raw} + \epsilon \mathcal{N}(0,1)$. We compute gradients through the environment during joint training to improve training performance.

\subsubsection{Results}
(See Appendix \ref{fig:rnnmdp_stoch} and \ref{fig:rnnmdp_gamma} for enlarged graphs). 

\begin{figure}[H]
  \caption{Varying Stochasticities for RNN-MDP}
  \label{fig:stoch_mdp}
  \centering
    \includegraphics[scale = 0.12]{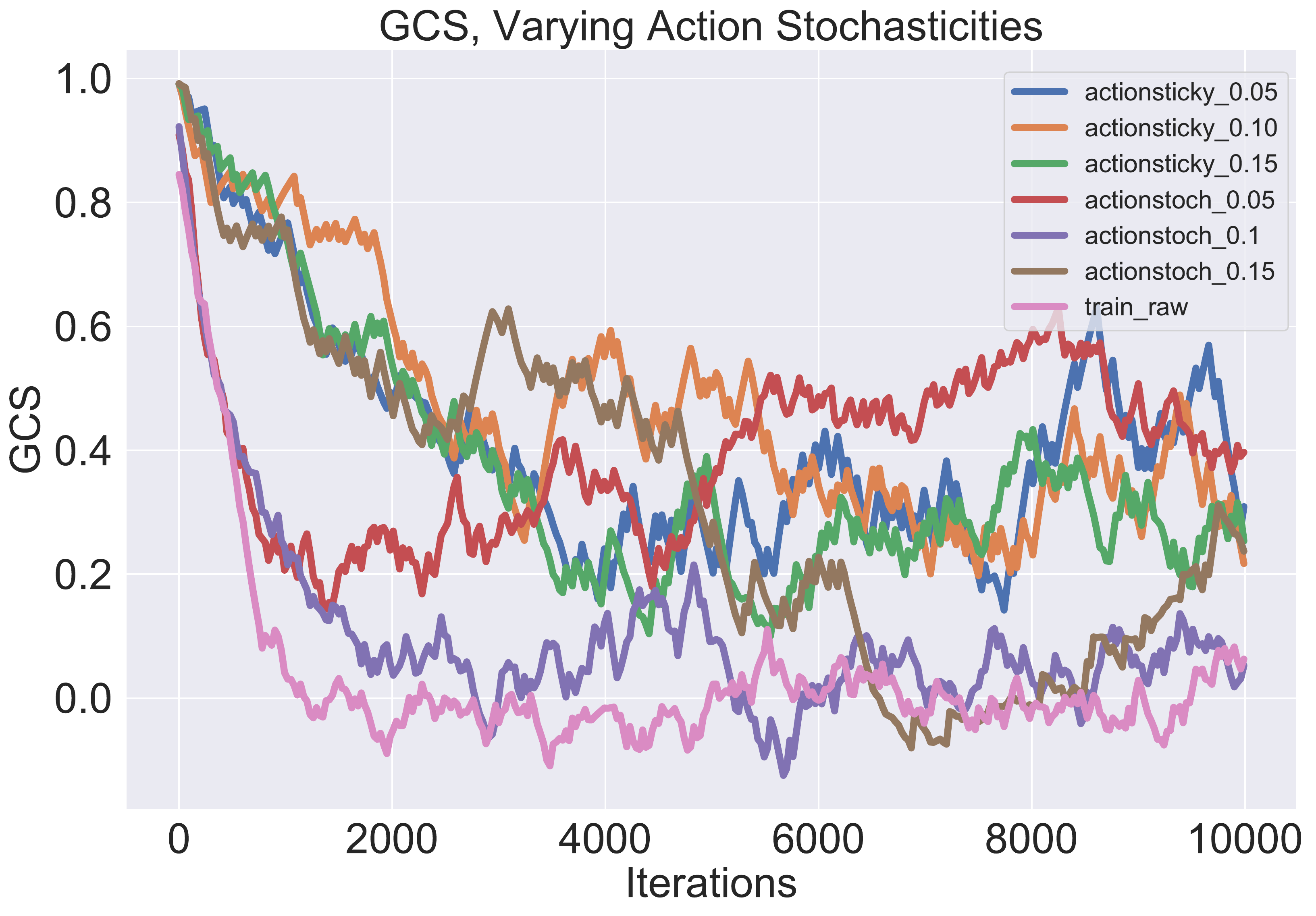}
    \includegraphics[scale = 0.12]{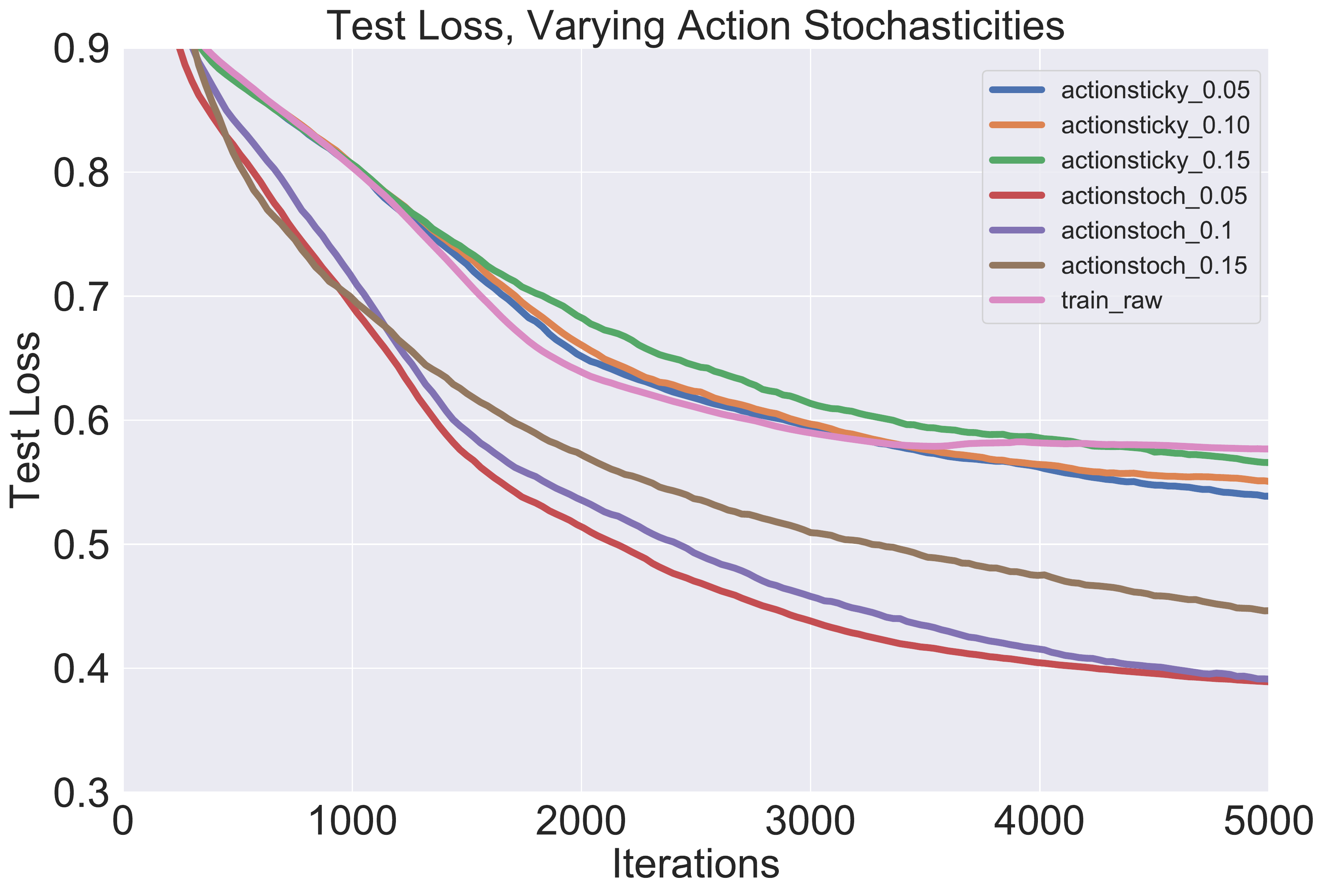}
\end{figure}

\begin{figure}[H]
  \caption{Varying $\gamma$ for nonstochastic and stochastic RNN-MDP}
   \label{fig:gamma_mdp}
  \centering
    \includegraphics[scale = 0.126]{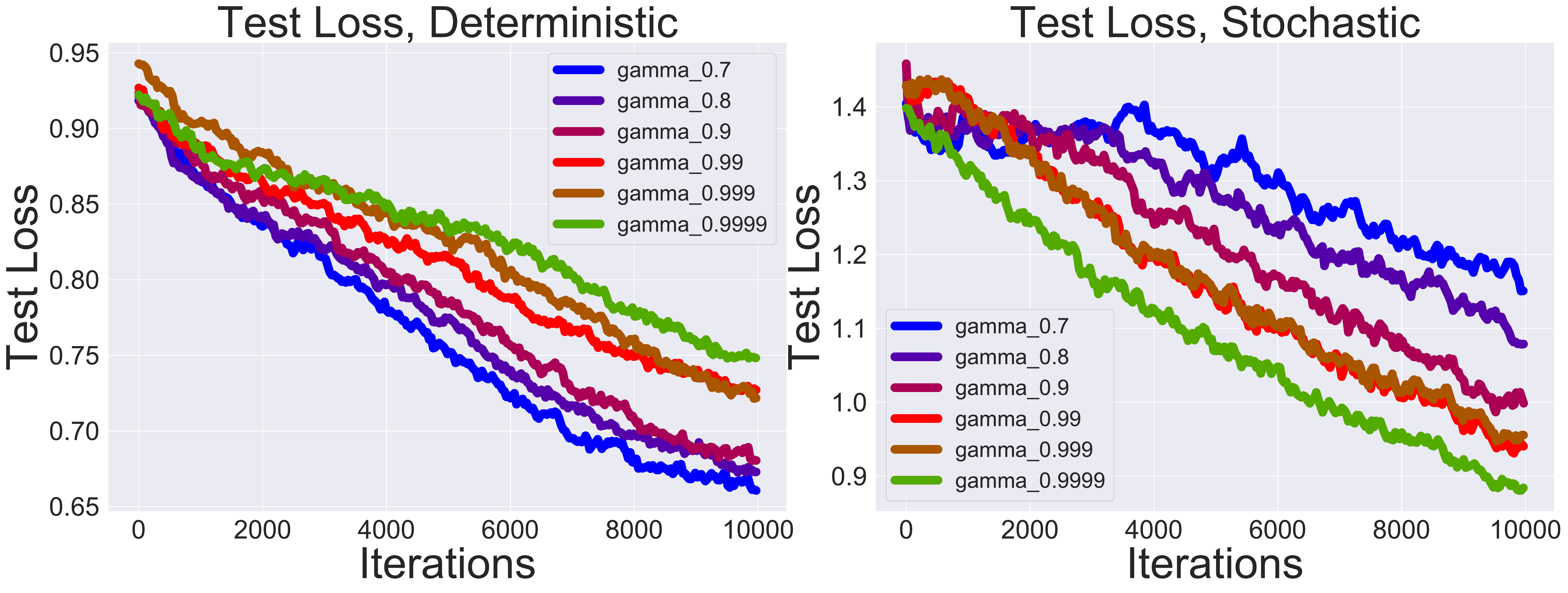}

\end{figure}

In \fig{fig:stoch_mdp}, we find adding stochasticity (e.g. random action stochasticity or sticky actions \cite{overfittingRL}) into actions increases GCS and decreases test loss. This result is isolated from any other factors in RL, and thus shows that stochasticity aligns the policy gradient more to the true gradient. Since entropy and stochasticity may smoothen the loss landscape in RL \cite{entropy_regularization}, this suggests that entropy may be also therefore be affecting the direction of the gradient in particular ways impacting generalization.

Furthermore, when introducing stochasticity into the environment by perturbing the state transition randomly at each turn, we find in \fig{fig:gamma_mdp} that higher gamma values in reward produces better generalization, while in the deterministic setting without transition stochasticity, lower gamma value produce better generalization. These seemingly conflicting results illustrates the complex interplay between action stochasticity and monotonicity of $\gamma$ with respect to generalization.

\subsection{Realistic MDP's - CoinRun}
\subsubsection{Differing Effects for Regularizers in RL }
\begin{figure}[H]  
  \caption{GCS vs Various Modifications (normal training, batchnorm, batchnorm with stochasticity = 0.1, stochasticity = 0.1, data augmentation)}
  \label{fig:coinrun_gcs}
  \centering
    \includegraphics[scale = 0.126]{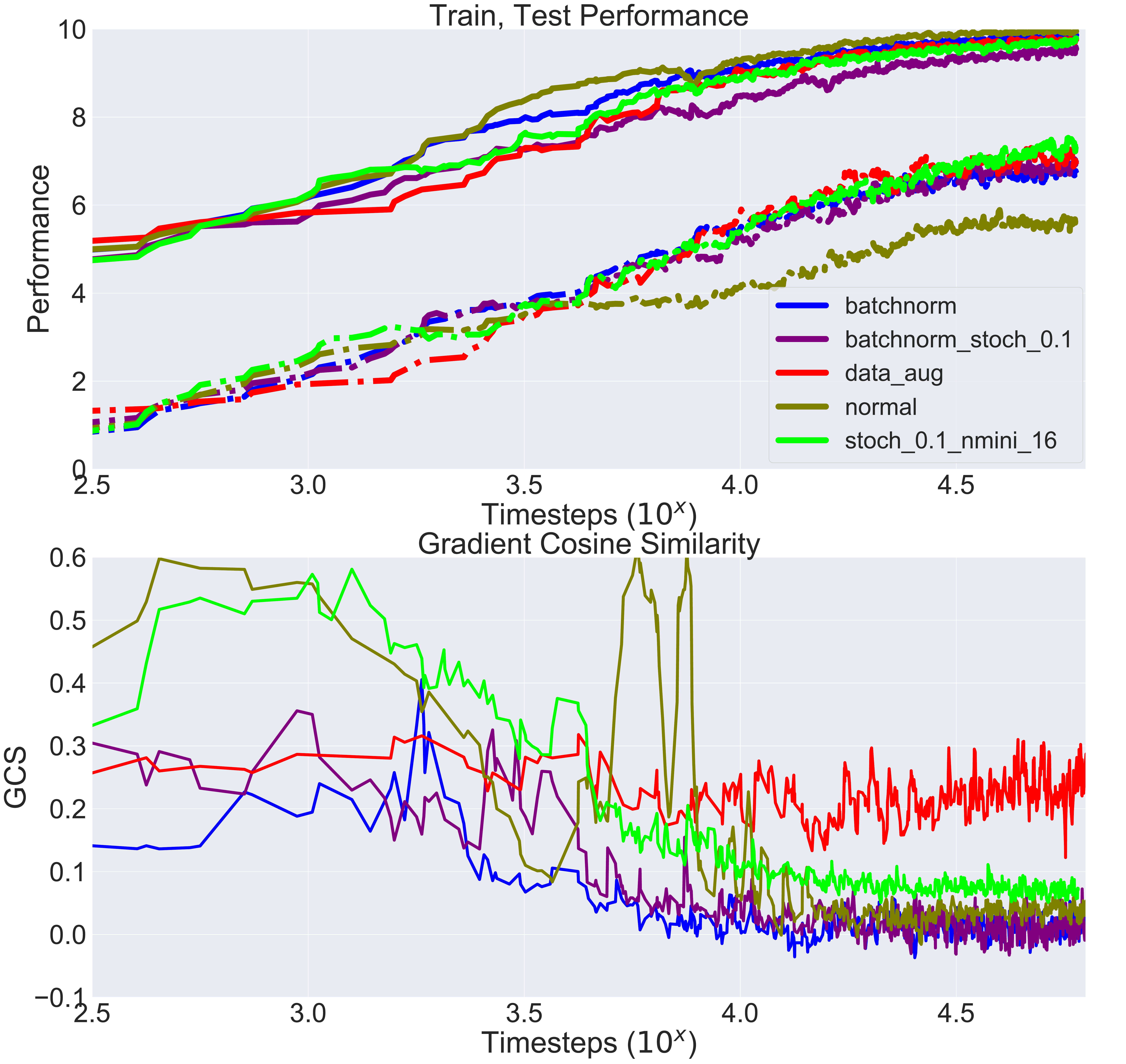}
    \label{fig:variousmods}
\end{figure}

We use CoinRun \cite{coinrun} as a procedurally generated MDP family for testing larger scale generalization and analyze common regularization techniques such as batch normalization, action stochasticity and data augmentation. Throughout all settings, in \fig{fig:coinrun_gcs}, we see that steadily decreasing gradient cosine similarity (GCS) produces overall worse testing performance, with higher asymptotic GCS correlated with test performance. However, regularizers have different effects: e.g. batchnorm changes the network weight norms while stochasticity does not, and stochasticity is heavily sensitive to batchsize.

In \fig{fig:coinrun_gcs}, we note that training can have instability on its gradient cosine similarity (GCS) and thus smoothness, but that such instability is also correated with poor test performance (see Appendix \ref{fig:gcs_unstable} for instability results).  

\begin{figure}[H]  
  \centering
        \caption{Trajectory Dependent Metrics vs Various Modifications (normal training, batchnorm, batchnorm with stochasticity = 0.1, stochasticity = 0.1, data augmentation)}
    \label{fig:variousmods_aux}
    \includegraphics[scale = 0.127]{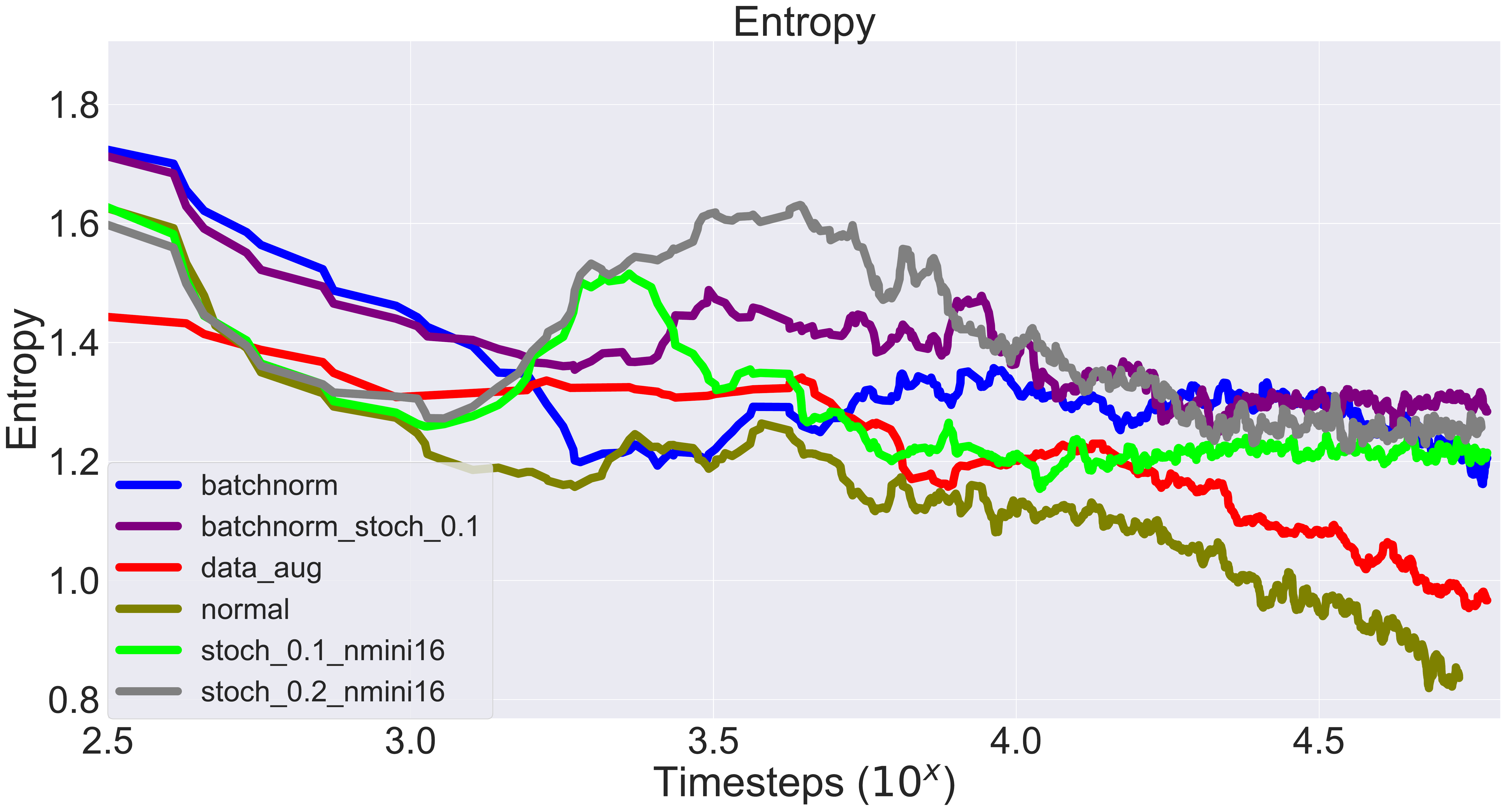}
    \includegraphics[scale = 0.127]{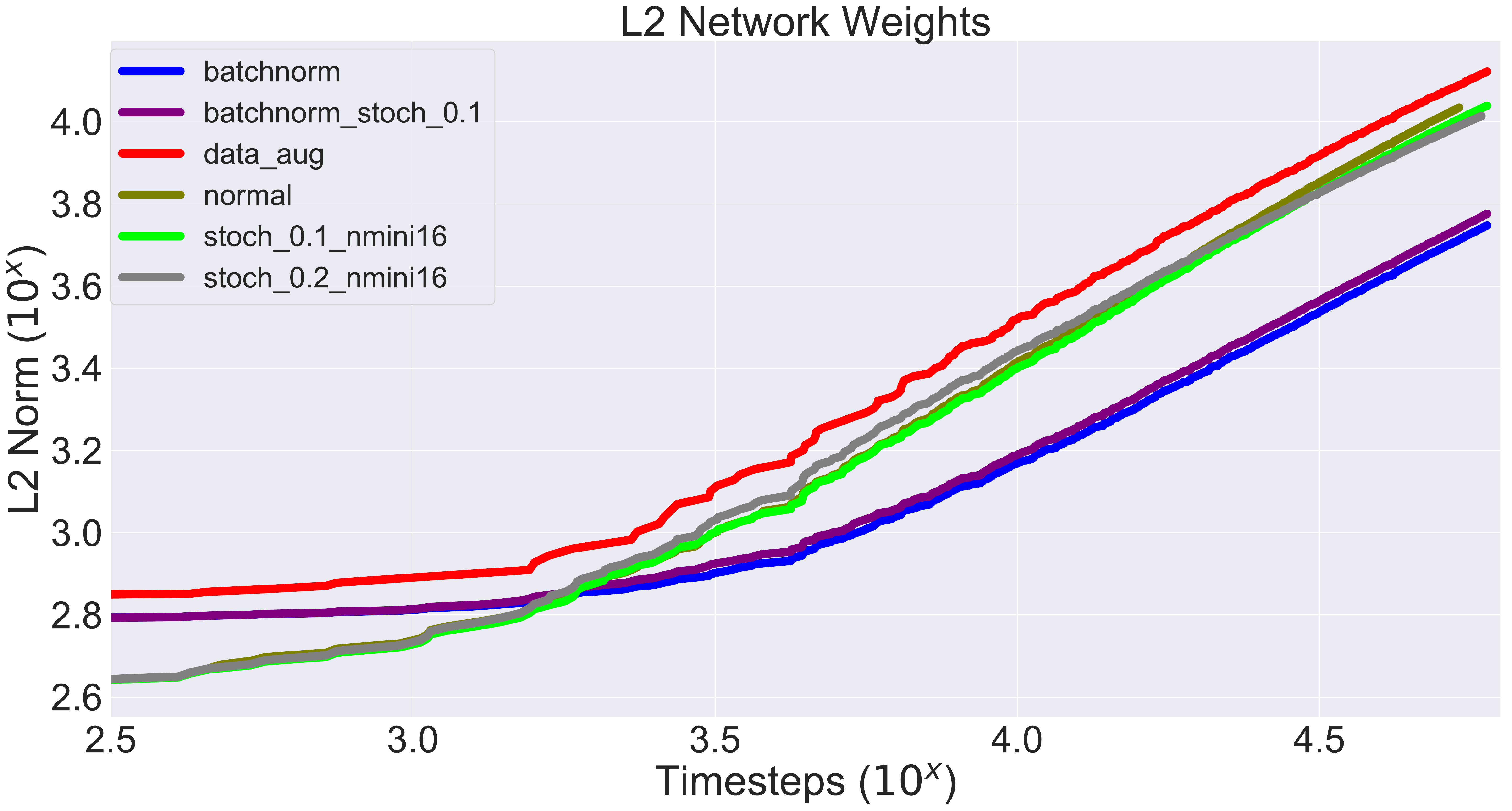}
    \includegraphics[scale = 0.127]{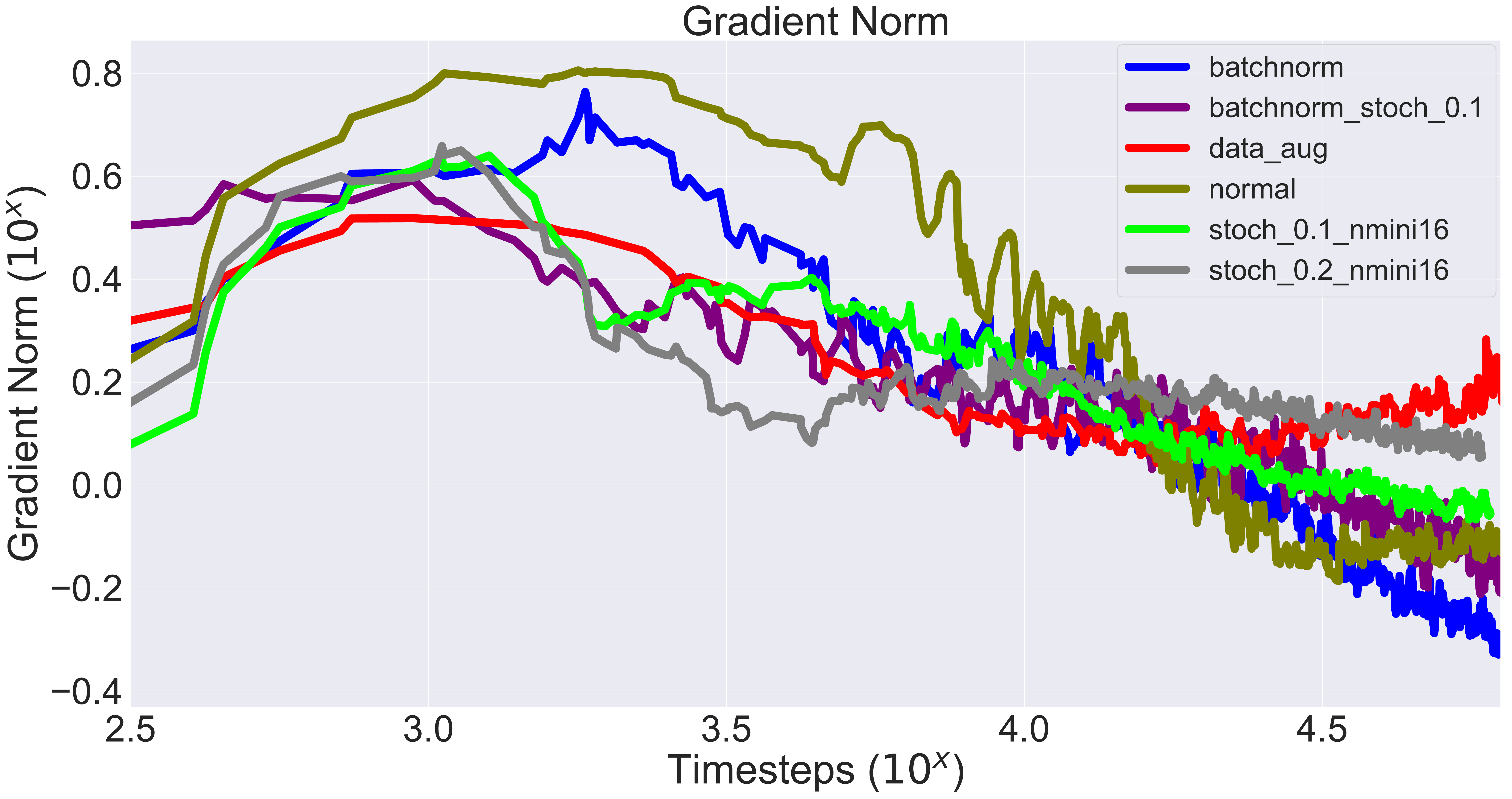}
\end{figure} 
In Figure \ref{fig:gcs_unstable}, \ref{fig:variousmods} we data augmentation produces a constant non-zero GCS, as well as high asymptotic gradient norm, but does not significantly deviate network norm or entropy. This implies that data augmentation has a strong effect on the Lipschitz constant on the loss, but also does not allow the policy to converge, which is consistent with the fact that its GCS does not converge to 0. 

Meanwhile, stochasticity leads to an overall better GCS consistent with \cite{entropy_regularization}, which correlates with the eventual higher test performance, consistent with the RNN-MDP environment. Batchnorm reduces both the variance and the magnitude of the GCS consistent with \cite{batchnorm_madry}, while the addition of stochasticity with batch normalization increases the GCS overall slightly. Furthermore, batchnorm also has a similar effect to stochasticity on the policy entropy, but stochasticity does not change the $\ell_{2}$ norm of network weights, while batchnorm does. Batchnorm reduces the gradient norm, as does stochasticity when compared based on reward vs gradient norm.

We encourage the reader to read Appendix \ref{fig:gcs_unstable}, \ref{fig:stoch_gradnorm}, which explores stochasticity's effects on gradient norm.

\subsubsection{Interdependence between Hyperparameters at a Larger Scale}
We investigate stochasticity's effects further, especially from batchsize. Both larger and smaller batchsizes generally \textit{benefit} batchnorm \cite{increasebatch}, and usually lower batchsizes improve generalization due to beneficial gradient noise \cite{largebatch} - we find that this is not the case for stochasticity. 

\begin{figure}[H] 
  \caption{GCS Instability with varying batchsize.}
  \centering
    \includegraphics[scale = 0.127]{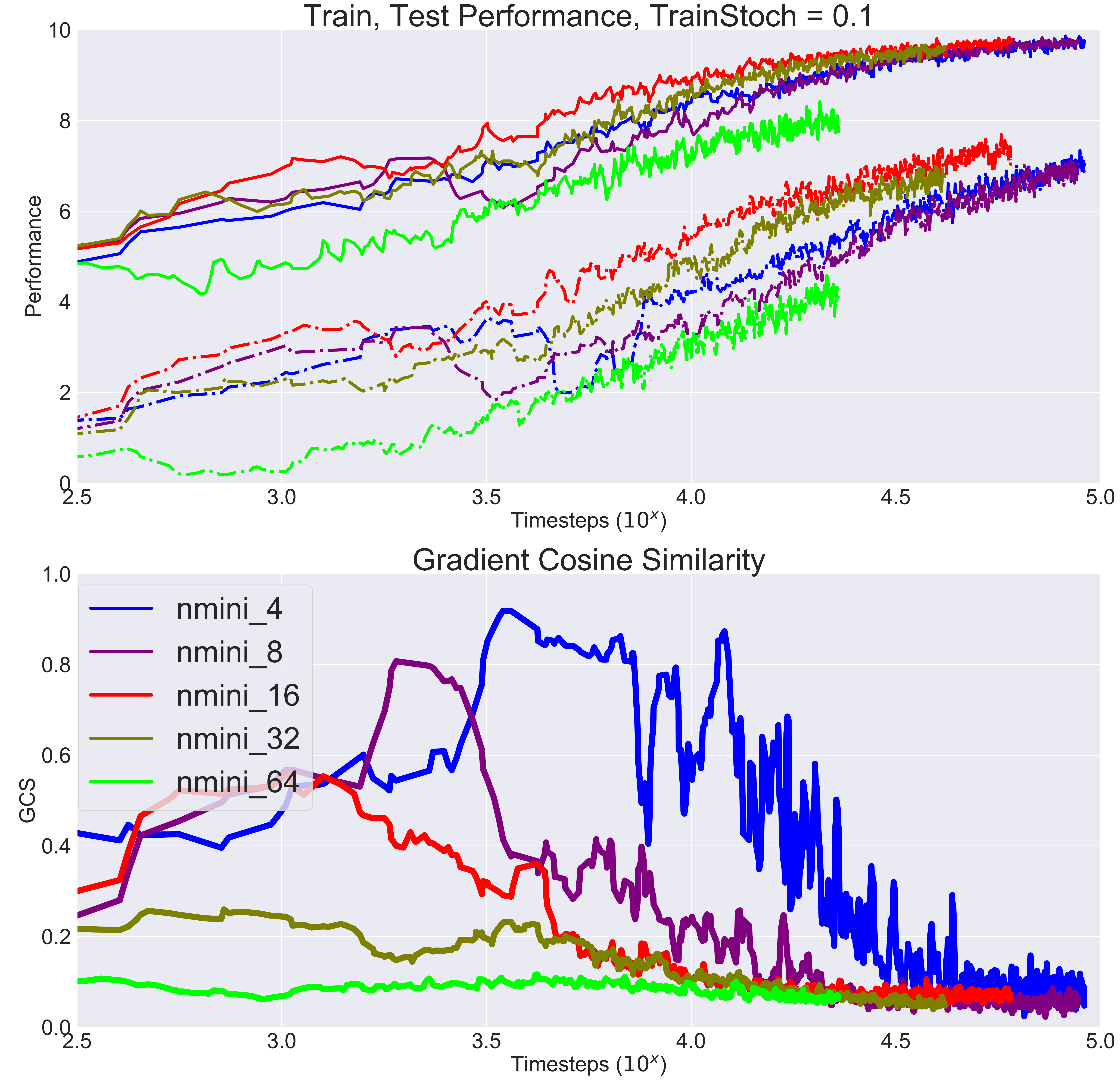}
    \label{fig:stochbatch}
\end{figure} 
\begin{figure}[H]
  \centering
    \includegraphics[scale = 0.127]{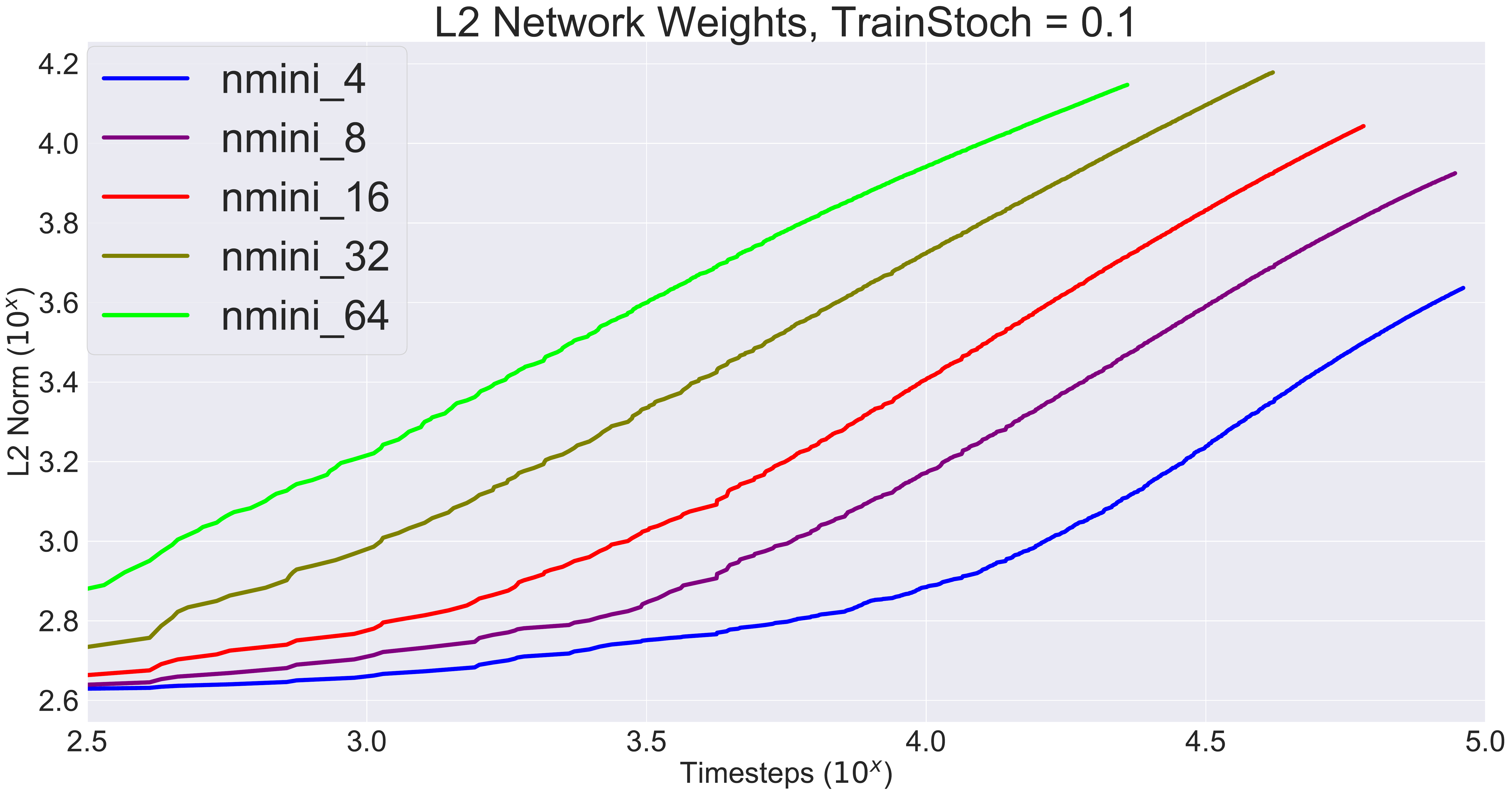}
\end{figure} 
As shown above, action stochasticity (entropy regularization) naturally boosts GCS - however, extra boosts can be caused by unwanted high variance on the GCS (which can be exacerbated by poor choices of batchsize, suggesting existence of sharp minimizer) and actually lead to poorer test performance. While keeping stochasticity fixed to 0.1 (Figure \ref{fig:stochbatch}) during training, the minibatch number (which inversely affects batchsize) has a strong effect on test performance and the weight norms, similar to SL. Surprisingly, higher batchsize (and thus less noise) produces significantly more variance on the GCS, also suggesting convergence on a sharp minimizer (i.e. high eigenvalue on the Hessian).

\begin{figure}[H] 
  \caption{Stochasticity vs NumMiniBatch} 
  \centering
    \includegraphics[scale = 0.13]{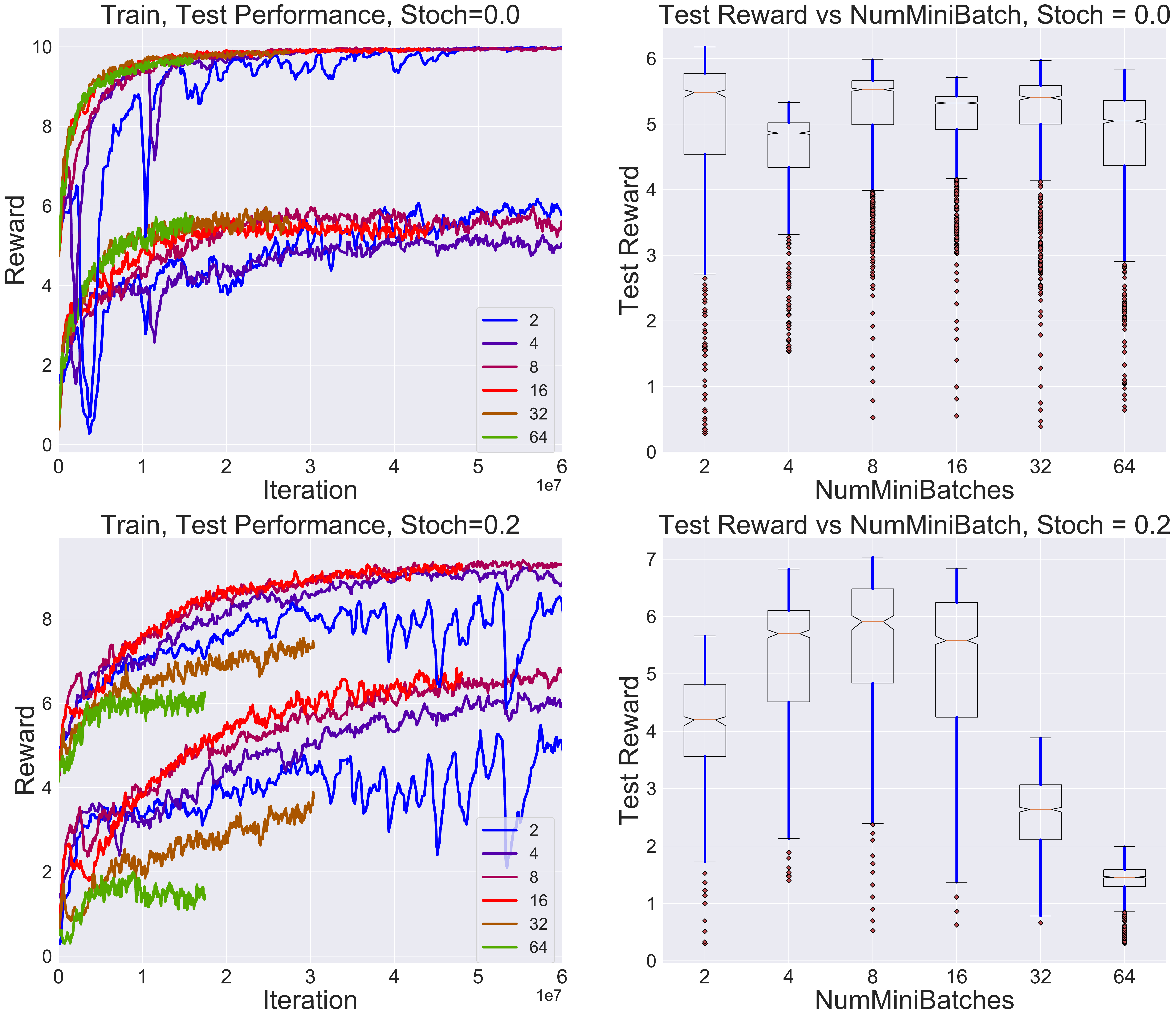}
    \label{fig:stoch_minibatch}
\end{figure}

Stochasticity also reduces the acceptable range of mini-batchnumber (Figure \ref{fig:stoch_minibatch}), showing this parameter is important to RL generalization when using entropy bonuses. The batchsize normally provides a form of gradient noise which may help training - in the RL case this is translated to sampling individual policy gradient estimates from a mixture between the true policy gradient and the entropy added gradient. As expected, too much noise caused by low batchsize such as nminibatch = 64 produces negligible GCS. 

We provide more relationship analysis in Appendix \ref{fig:gammastoch_full}, testing relationships between stochasticity and $\gamma$ on CoinRun and CoinRun-Mazes (an RNN exploration test).

\section{Conclusion}
From the above results, we conclude that when dealing with RL generalization, one must be very careful in hyperparameter optimization. As shown in this work, hyperparameters are simply not orthogonal in terms of improving generalization performance. One change in a hyperparameter may change the monotonicity relationship for another hyperparameter, and thus and may easily provide wrong conclusions during experimentation. In comparison to SL, RL also has many more important hyperparameters, all of which must be considered carefully when designing robust systems.

There is a lack of understanding on these relationships, and a deeper investigation could prove important for both the practice and theory of RL-generalization. We hope that this work provides guidance for both future work and practical guidance in designing robust and generalizable RL systems.

\clearpage
%\nocite{*}
\bibliographystyle{icml2019}
\bibliography{references}

\begin{thebibliography}{13}
\providecommand{\natexlab}[1]{#1}
\providecommand{\url}[1]{\texttt{#1}}
\expandafter\ifx\csname urlstyle\endcsname\relax
  \providecommand{\doi}[1]{doi: #1}\else
  \providecommand{\doi}{doi: \begingroup \urlstyle{rm}\Url}\fi

\bibitem[Ahmed et~al.(2018)Ahmed, Roux, Norouzi, and
  Schuurmans]{entropy_regularization}
Ahmed, Z., Roux, N.~L., Norouzi, M., and Schuurmans, D.
\newblock Understanding the impact of entropy on policy optimization.
\newblock \emph{CoRR}, abs/1811.11214, 2018.

\bibitem[Cobbe et~al.(2018)Cobbe, Klimov, Hesse, Kim, and Schulman]{coinrun}
Cobbe, K., Klimov, O., Hesse, C., Kim, T., and Schulman, J.
\newblock Quantifying generalization in reinforcement learning.
\newblock \emph{CoRR}, abs/1812.02341, 2018.

\bibitem[Henderson et~al.(2018)Henderson, Islam, Bachman, Pineau, Precup, and
  Meger]{deeprl_matters}
Henderson, P., Islam, R., Bachman, P., Pineau, J., Precup, D., and Meger, D.
\newblock Deep reinforcement learning that matters.
\newblock In \emph{Proceedings of the Thirty-Second {AAAI} Conference on
  Artificial Intelligence, (AAAI-18), the 30th innovative Applications of
  Artificial Intelligence (IAAI-18), and the 8th {AAAI} Symposium on
  Educational Advances in Artificial Intelligence (EAAI-18), New Orleans,
  Louisiana, USA, February 2-7, 2018}, pp.\  3207--3214, 2018.
\newblock URL
  \url{https://www.aaai.org/ocs/index.php/AAAI/AAAI18/paper/view/16669}.

\bibitem[Ilyas et~al.(2018)Ilyas, Engstrom, Santurkar, Tsipras, Janoos,
  Rudolph, and Madry]{madrypolicygradient}
Ilyas, A., Engstrom, L., Santurkar, S., Tsipras, D., Janoos, F., Rudolph, L.,
  and Madry, A.
\newblock Are deep policy gradient algorithms truly policy gradient algorithms?
\newblock \emph{CoRR}, abs/1811.02553, 2018.

\bibitem[Ioffe \& Szegedy(2015)Ioffe and Szegedy]{batchnorm}
Ioffe, S. and Szegedy, C.
\newblock Batch normalization: Accelerating deep network training by reducing
  internal covariate shift.
\newblock In \emph{{ICML}}, volume~37 of \emph{{JMLR} Workshop and Conference
  Proceedings}, pp.\  448--456. JMLR.org, 2015.

\bibitem[Jiang et~al.(2015)Jiang, Kulesza, Singh, and Lewis]{dependencehorizon}
Jiang, N., Kulesza, A., Singh, S., and Lewis, R.
\newblock The dependence of effective planning horizon on model accuracy.
\newblock In \emph{Proceedings of the 2015 International Conference on
  Autonomous Agents and Multiagent Systems}, AAMAS '15, pp.\  1181--1189,
  Richland, SC, 2015. International Foundation for Autonomous Agents and
  Multiagent Systems.
\newblock ISBN 978-1-4503-3413-6.

\bibitem[Keskar et~al.(2016)Keskar, Mudigere, Nocedal, Smelyanskiy, and
  Tang]{largebatch}
Keskar, N.~S., Mudigere, D., Nocedal, J., Smelyanskiy, M., and Tang, P. T.~P.
\newblock On large-batch training for deep learning: Generalization gap and
  sharp minima.
\newblock \emph{CoRR}, abs/1609.04836, 2016.

\bibitem[Li et~al.(2018)Li, Xu, Taylor, Studer, and Goldstein]{visualizingloss}
Li, H., Xu, Z., Taylor, G., Studer, C., and Goldstein, T.
\newblock Visualizing the loss landscape of neural nets.
\newblock In \emph{Advances in Neural Information Processing Systems 31: Annual
  Conference on Neural Information Processing Systems 2018, NeurIPS 2018, 3-8
  December 2018, Montr{\'{e}}al, Canada.}, pp.\  6391--6401, 2018.

\bibitem[Santurkar et~al.(2018)Santurkar, Tsipras, Ilyas, and
  Madry]{batchnorm_madry}
Santurkar, S., Tsipras, D., Ilyas, A., and Madry, A.
\newblock How does batch normalization help optimization?
\newblock In Bengio, S., Wallach, H., Larochelle, H., Grauman, K.,
  Cesa-Bianchi, N., and Garnett, R. (eds.), \emph{Advances in Neural
  Information Processing Systems 31}, pp.\  2487--2497. Curran Associates,
  Inc., 2018.

\bibitem[Schulman et~al.(2017)Schulman, Wolski, Dhariwal, Radford, and
  Klimov]{ppo}
Schulman, J., Wolski, F., Dhariwal, P., Radford, A., and Klimov, O.
\newblock Proximal policy optimization algorithms.
\newblock \emph{CoRR}, abs/1707.06347, 2017.

\bibitem[Smith et~al.(2017)Smith, Kindermans, and Le]{increasebatch}
Smith, S.~L., Kindermans, P., and Le, Q.~V.
\newblock Don't decay the learning rate, increase the batch size.
\newblock \emph{CoRR}, abs/1711.00489, 2017.

\bibitem[Williams(1992)]{reinforce}
Williams, R.~J.
\newblock Simple statistical gradient-following algorithms for connectionist
  reinforcement learning.
\newblock \emph{Machine Learning}, 8:\penalty0 229--256, 1992.
\newblock \doi{10.1007/BF00992696}.

\bibitem[Zhang et~al.(2018)Zhang, Vinyals, Munos, and Bengio]{overfittingRL}
Zhang, C., Vinyals, O., Munos, R., and Bengio, S.
\newblock A study on overfitting in deep reinforcement learning.
\newblock \emph{CoRR}, abs/1804.06893, 2018.

\end{thebibliography}

\onecolumn
\appendix
\renewcommand{\thesection}{A.\arabic{section}}
\renewcommand{\thefigure}{A\arabic{figure}}
\setcounter{section}{0}
\setcounter{figure}{0}
\title{Appendix: Understanding Generalization in Reinforcement Learning with a Supervised Learning Framework}
\date{}
\maketitle

\section{Full Results, Enlarged Pictures, Examples for Completeness}

\subsection{Why the RNN-MDP?}
For the sake of simplicity, assume our observation was the identity function (i.e. we observe the state). Note that the "state" within an MDP is subjective, depending on certain modifications, and this can dramatically affect solvability. For instance, an agent whose observation only consists of single timestep states is unable to adapt to different transition functions in different MDP's. This can occur in Mujoco when two MDP's possesss different gravities, and thus the optimal action is different even if the witnessed states are the same. However, an addition of framestacking will change the "witnessed state" to rather, a combination of last 4 timesteps from which an agent will be able to adapt to different gravities. Thus, we cannot simply generate "random-MDP"s as a valid benchmark.

Thus, to prevent these ambiguities, we simply construct a family of MDP's by using one single transition $\mathcal{T}(s_{t}, a_{t}) \rightarrow s_{t+1}$, but use different initial starting states, and let reward $r$ be independent of $\theta$, and $\theta$ controls only the initial state $s_{0}$. This solves the above problem if the state also contains the gravity parameter. This way, there is always one optimal action given each state. We use a non-linear RNN cell (with additional stochasticities as an option) to simulate the transition $\mathcal{T}$, as well as nonlinear functions for reward. Conceptually, an RNN may approximate any sequential transition, and thus with a large enough cell, may simulate any MDP.

\subsection{RNN MDP Graphical Results}
We see that various forms of stochasticity (pure random stochasticity as well as action stickyness, with probabilities shown on legend) also produce higher GCS, as well as better testing performance.
\begin{figure}[H]
  \caption{Varying Stochasticities for RNN-MDP}
  \centering
    \includegraphics[scale = 0.25]{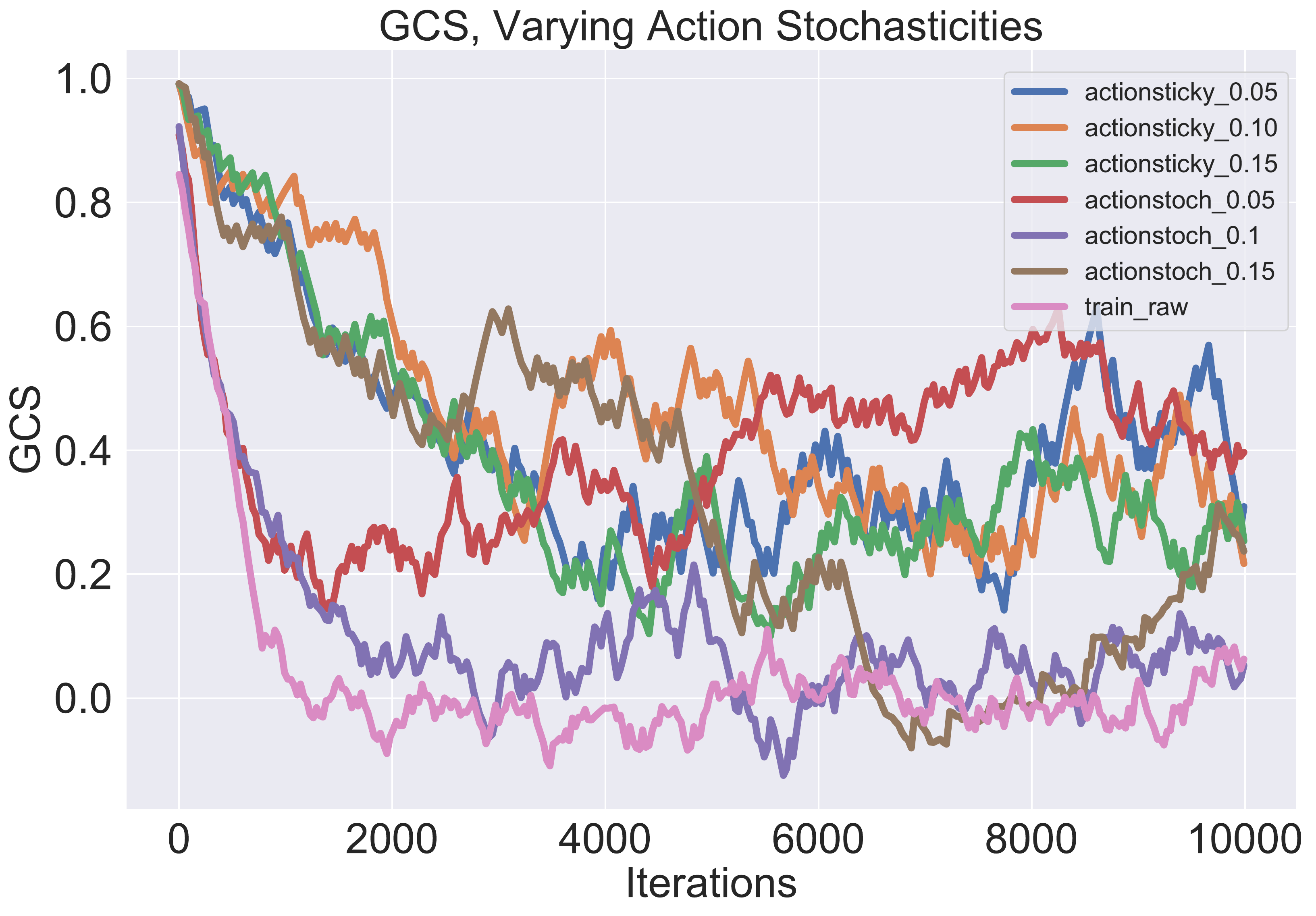}
    \includegraphics[scale = 0.25]{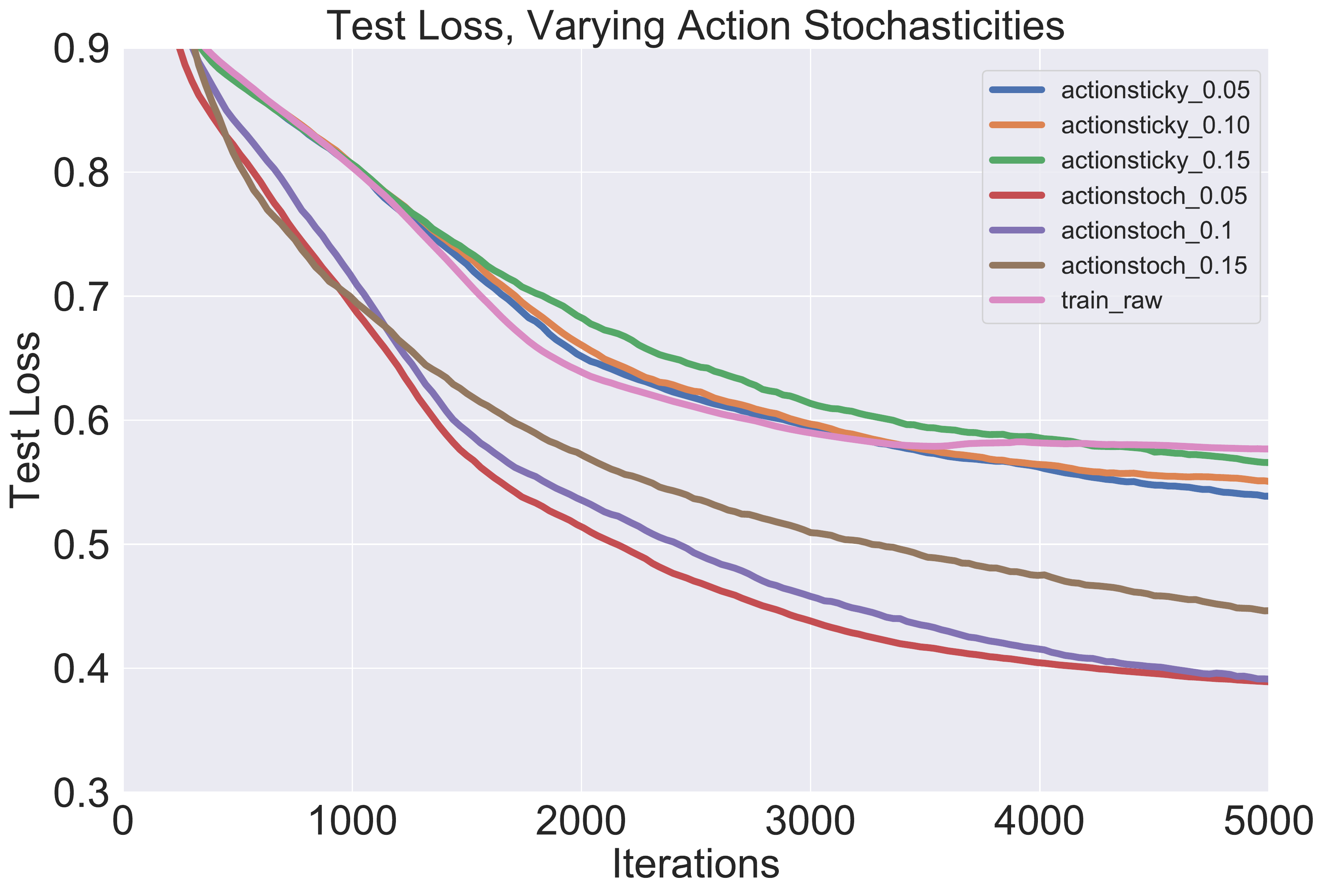}
    \label{fig:rnnmdp_stoch}
\end{figure}

\subsubsection{$\gamma$ in RNN-MDP}
We apply a similar stochastic modification on the RNN-MDP, by adding noise into the state: $s^{true}_{t} = s_{t} + \mathcal{N}(0.0, \sigma)$ and find that stochasticity in the environment produces a monotone relationship in which higher $\gamma$ produced better test performance, while determinism in the environment reverses this relationship - lower $\gamma$ produces better testing performance.
\begin{figure}[H]
  \caption{Varying $\gamma$ for nonstochastic and stochastic RNN-MDP}
  \centering
    \includegraphics[scale = 0.25]{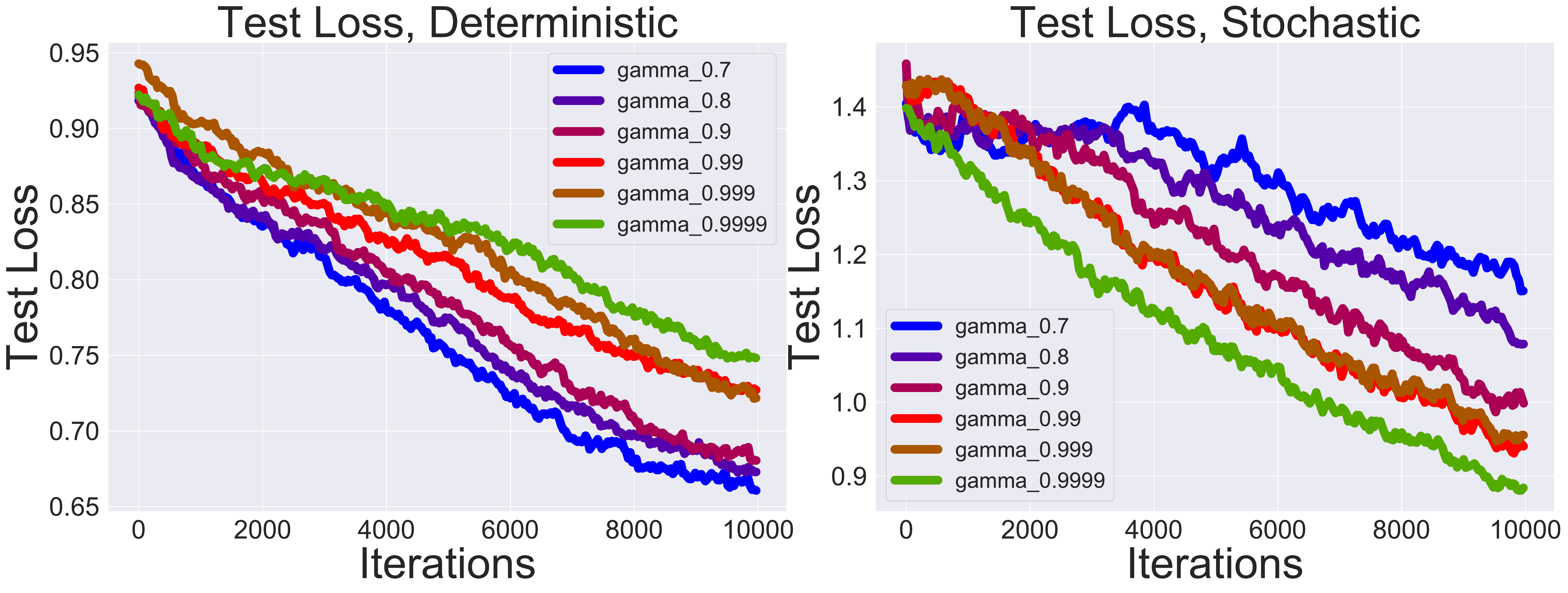}
    \label{fig:rnnmdp_gamma}
\end{figure}

\subsection{Why CoinRun?}
\begin{figure}[H]
  \caption{Varying $\gamma$ for nonstochastic and stochastic mazes.} 
  \centering
    \includegraphics[scale = 0.26]{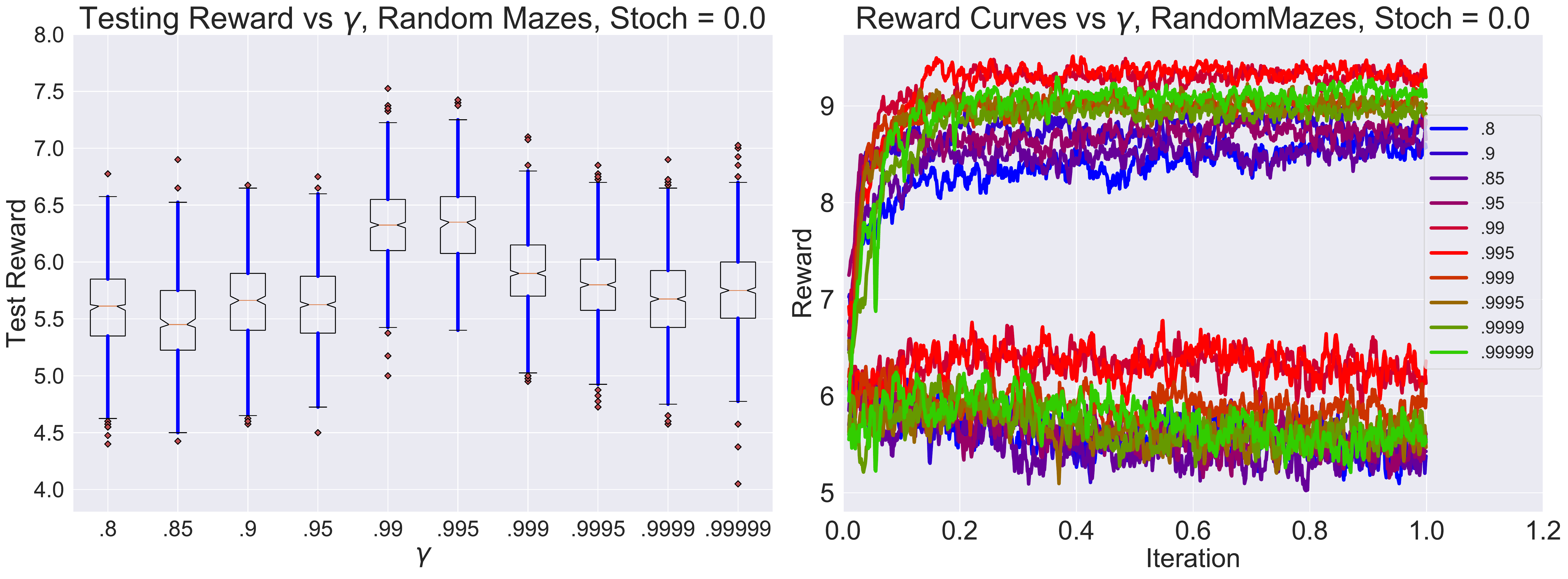}
    \includegraphics[scale = 0.26]{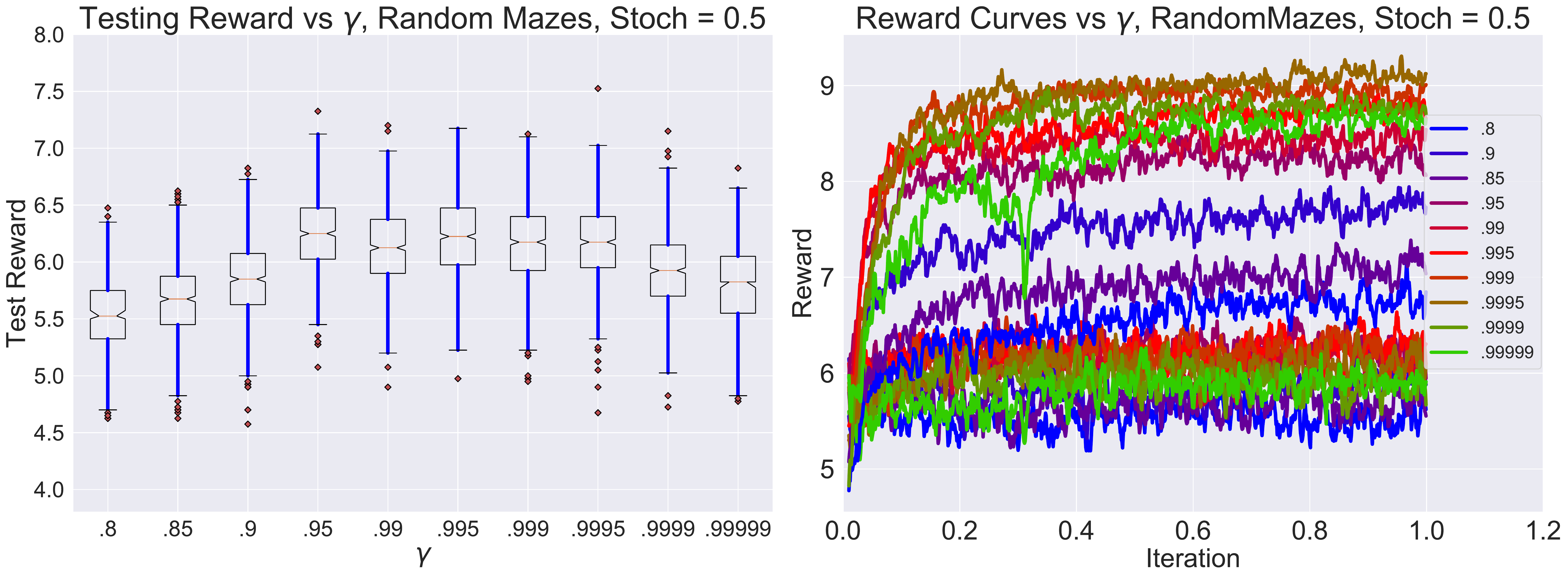}
    \label{fig:maze}
\end{figure}

Coinrun Package (Coinrun-Standard, RandomMazes) \cite{coinrun} - Both environments provide infinite-level-generation; Coinrun-Standard is a sidescroller which particularly benchmarks strong convolutional network generalization and establishes various regularizations indeed help with test performance. RandomMazes (in \ref{fig:maze} are sequentially generated mazes which test the exploration properties of RNN's. 

\subsection{Regularizers in RL}
\label{appendix:caveat}
As a reminder to the reader, it should be noted that all algorithms produce a $\pi_{\phi}$ that optimizes a surrogate loss, but at test time, is evaluated on $\E_{\theta \sim \Theta} \left[R_{\theta}(\pi_{\phi}) \right]$. For example, adding the discount factor $\gamma$ and entropy regularization make the surrogate training objective \begin{equation} R^{surrogate}(\pi_{\phi}) =  \left( S(\pi_{\phi}) + \sum_{t = 1}^{T}  \gamma^{i} r_{t} \right) \end{equation}

We formally introduce the exact regularizers and what their effects may entail:
\begin{compactitem}
\item Action Stochasticity:  \cite{entropy_regularization} argues that entropy bonuses and action stochasticity provides a smoother loss function. In the generalization case, we study its biasing effects on aligning the training set policy gradient with the true distribution policy gradient. In $T = 1$ case, $a_{1}$ is generated from $p_{base}$ with probability $\alpha$, which can be reinterpreted as purely adding $KL(p_{base} || \pi_{\beta})$ to the objective. The difference in gradient cosine is examining the contribution of $\nabla_{\beta} KL(p_{base} || \pi_{\beta}) \cdot \nabla_{\beta} R(\pi_{\beta})$.

We provide the normal policy gradient as found in \cite{reinforce} where $\tau = (s_{1},a_{1},...,s_{T}, a_{T})$:
$$ \nabla_{\beta} J(\beta) = \E_{\tau} \left[ \left( \sum_{t=1}^{T} r_{t} \right) \left(\sum_{t=1}^{T} \nabla_{\beta}  \log  \pi_{\beta}(a_{t} | s_{t}) \right) \right] $$

Suppose for the sake of simplicity we considered the $T = 1$ horizon case, and used action stochasticity, where $a_{1}$ is generated from $p_{base}$ with probability $\alpha$; thus the policy gradient for a single batch instance is a mixture, i.e. $\widehat{\nabla} J_{s \in \mathcal{D}_{s}}(\beta)$ is 

$ \E_{a \sim p_{base} } r(s, a) \nabla_{\beta} \log \pi (a | s)$ with probability $\alpha$ and $\E_{a \sim \pi(a | s) } r(s, a) \nabla_{\beta} \log \pi (a | s)$ with probability $(1-\alpha)$.
This implies that in the infinite batch setting, we have: 
$$\E_{s \in \mathcal{D}_{s}}\left[ \nabla J_{s}(\beta) \right] = \E_{s \in \mathcal{D}_{s}} \left[ \alpha \E_{a_{1} \sim p_{base} } \left[ r(s, a_{1}) \nabla_{\beta} \log \pi_{\beta} (a_{1} | s) \right]+  (1-\alpha)\E_{a_{2} \sim \pi(a | s) } \left[r(s, a_{2}) \nabla_{\beta} \log \pi_{\beta} (a_{2} | s) \right] \right] $$ 

Note that $$\E_{a \sim p_{base} } \left[ r(s, a) \nabla_{\beta} \log \pi_{\beta} (a| s) \right] = \nabla_{\beta} \left[ \int_{a} r(s,a) p_{base}(a) \log \pi_{\beta}(a | s) da \right]$$ with cross entropy $H(p_{base}, \pi_{\beta}) =  -\int_{a} p_{base}(a) \log \pi_{\beta}(a | s) da$. Thus we may view this as a reward-weighted entropy penalty: if $r(s,a)$ was a relatively uniform distribution, then we are simply taking the gradient respect to the KL between the policy and a uniform distribution: $\E_{a \sim p_{base} } \left[ r(s, a) \nabla_{\beta} \log \pi_{\beta} (a| s) \right] \approx -\nabla_{\beta} KL(p_{base} || \pi_{\beta})$.

As an aside, for a discrete softmax policy, $\nabla \log \pi(a | s) = ( 1 - \pi_{\beta}(a | s)) \nabla_{\beta} f_{\beta}(s, a) $, and hence the contribution from the random action will be:

$$\E_{a \sim p_{base} } \left[ r(s, a) \nabla_{\beta} \log \pi (a | s) \right] =  \E_{a \sim Uniform } \left[ r(s, a)  ( 1 - \pi_{\beta}(a | s)) \nabla_{\beta} f_{\beta}(s, a) \right]$$

At a high level, as the policy's entropy becomes lower and hence the policy is more confident, both the term $(1- \pi_{\beta}(a | s))$ and the norm of the gradient $\nabla_{\beta} f_{\beta}(s,a)$ become high.

We see that using an entropy penalty ${KL}(p_{base} || \pi_{\beta})$ to the objective at only \textit{training} time will align better with the true policy gradient. Unlike linear regression where adding an explicit regularizer on the weight norm decreases the weights during training, this is not the case for RL, as we see that stochasticity does not provide a statistically significant change to the $\ell_{2}$ norm of the network weights. 

Note that this noise addition is not unbiased in parameter space - if adding a simple random Gaussian noise to a gradient $g_{t}' = g_{t} + \mathcal{N}(0,\sigma)$, then on expectation with respect to the noise sample, the expectation of the dot product between training sample gradient $\widehat{g}_{t}$ and true gradient $g_{t}^{true}$ remains fixed: 
$$ \E_{\epsilon_{1}, \epsilon_{2} \in \mathcal{N}(0, \sigma)} \left[(\widehat{g}_{t} + \epsilon_{1})(g_{t}^{true} + \epsilon_{2}) \right] = \widehat{g}_{t} \cdot g_{t}^{true} $$

\item BatchNorm \cite{batchnorm}: \cite{coinrun} empirically showed the benefit of batchnorm on RL generalization. Translated to the RL-setting, \cite{batchnorm_madry} establishes that for every action $a$, if $f^{BN}$ is after adding batchnorm on $f$, batchnorm can reduce the norm on the gradient as well as the 2nd order Hessian (smoothness) term: $\norm{\nabla_{\beta} f^{BN}(s,a)}_{2}^{2} \le \norm{\nabla_{\beta}f(s,a)}_{2}^{2}$ and $\nabla_{\beta} f^{BN}(s,a)^{T} H_{aa}^{f_{BN}} \nabla_{\beta} f^{BN}(s,a) \le \nabla_{\beta} f(s,a)^{T} H_{aa}^{f} \nabla_{\beta} f(s,a) $. We thus expect its gradient smoothing effects also reduce the variance on the gradient cosine similarity, as well as provide a smoother transition from the beginning of training to the end. 

We see that \cite{batchnorm_madry}'s results on smoothing both the gradient norm and the Hessian term translate to the policy gradient. They state that for a network $f^{BN}$ which uses batchnorm layers while $f$ is the original network, for a loss function $\mathcal{L}(f_{\beta})$ (shortened to $\mathcal{L}$, denote $\mathcal{L}_{BN}$ to be the loss function after adding batch norm:
\begin{equation}
\norm{\nabla_{a} \mathcal{L}_{BN}}^{2} \le \frac{\phi^{2}}{\sigma^{2}_{a}} \left( \norm{\nabla_{a} \mathcal{L}}^{2} - \frac{1}{m} \langle \mathbf{1}, \nabla_{a}\mathcal{L} \rangle  - \frac{1}{\sqrt{m}} \langle \nabla_{a} \mathcal{L}, \widehat{a} \rangle^{2} \right)
\end{equation}
and Hessian smoothness term

\begin{equation}
\nabla_{a} \mathcal{L}_{BN}^{T} \frac{\partial \mathcal{L}_{BN}}{\partial a \partial a}  \nabla_{a} \mathcal{L}_{BN} \le \frac{\phi^{2}}{\sigma^{2}_{a}} \left( \nabla_{a} \mathcal{L}^{T} H_{jj} \nabla_{a} \mathcal{L}  - \frac{1}{m \phi } \langle \nabla_{a} \mathcal{L}_{BN}, \widehat{a} \rangle  \norm{ \frac{\partial \mathcal{L}_{BN}}{\partial a} }^{2} \right)
\end{equation}

where $a$ is an activation, $\widehat{a}$ is the set of activations after batch norm, $m$ is batchsize, $\sigma_{a}$ is the standard deviation computed over the minibatch of $a$'s, $\phi$ is a constant term. In reinforcement learning, $\mathcal{L}(f_{\beta})$ is thus $\E_{\theta \sim \Theta} \left[ R_{\theta}(f_{\beta}) \right]$. The contributions to the smoothness come from the terms to the right in both equations, which are subtracting off from the gradient norm and Hessian term respectively of the original network. Thus from the above, increasing batchsize $m$ generally reduces batchnorm's smoothing effects.

Since experimental results show that entropy also increased asymptotically, this implies that $(1- \pi_{\beta}( a| s))$ contributes to the gradient norm significantly, while the network gradient $\nabla_{\beta} f_{\beta}(s,a)$ norm is low, consistent with the lower $\ell_{2}$ norm during training.

\item $\gamma$ - \cite{dependencehorizon} shows that higher $\gamma$ increases the size of the optimal policy set for the training MDP's, because $\gamma$ will force the policy space to consider multiple state-action pairs for planning. Thus $\gamma$ can also be a useful tool to verify the number of local minima of the training landscape. If high $\gamma$ produces worse \textit{testing} performance, this implies that the set of optimal solutions for the true distribution is small, and the agent has overfitted by converging on one of the many optimal policies for the \textit{training set}.

\item For data augmentation, a reasonable model is to sample $s' \sim p_{aug}( \cdot | s)$, where $p_{aug}$ randomly adds shapes to the picture, thus our policy gradient is \begin{equation} \E_{s' \sim p_{aug}(\cdot | s), s \sim \mathcal{D}_{s}, a \in \pi(a | s')} \left[r(s,a) \nabla_{\beta} \log \pi_{\beta}(a|s')\right] \end{equation}

If we consider for simplicity, a deterministic function $w(\cdot): s \rightarrow s'$, then the gradient is instead $\nabla \log \pi(a | s') = (1- \pi_{\beta}(a| w(s))) \nabla_{\beta}f_{\beta}(w(s), a)$. As the entropy is not significantly raised at convergence from experimental results, a mild assumption is $(1- \pi_{\beta}(a | w(s)) \approx (1- \pi_{\beta}(a | s))$ and thus $\nabla_{\beta}f_{\beta}(w(s),a)$ has strong alignment with $\nabla_{\beta}f_{\beta}(s,a)$.

\end{compactitem}

\newpage 
\subsection{Extended CoinRun Results}
\subsubsection{GCS's Correlation with Test Instability}
Runs with roughly non-monotonically decreasing GCS tended to also produce more unstable \textit{testing} curves, even with stable training curves. These large "bumps" in the GCS corresponded to sudden drops on the testing curves. Furthermore, testing curves that produced such "bumps" ultimately produced poorer final testing performances. In the monotonically decreasing GCS cases, this corresponded to better testing performance overall.

In terms of loss landscape, when both training and testing performance drop, this implies that the trajectory of gradient descent has accidentally reached a peak on the true distribution's landscape, from too large of a step size. However, other times exist in which the GCS sharply rises when the training curve is stable while testing curve is not, suggesting the lack of smoothness on training landscape is causing instability.
\begin{figure}[H]
  \caption{GCS's noisy "bumps" can match with "bumps" in training performance. (GCS y-axis is scaled by y/10.0.)} 
  \centering
    \includegraphics[scale = 0.26]{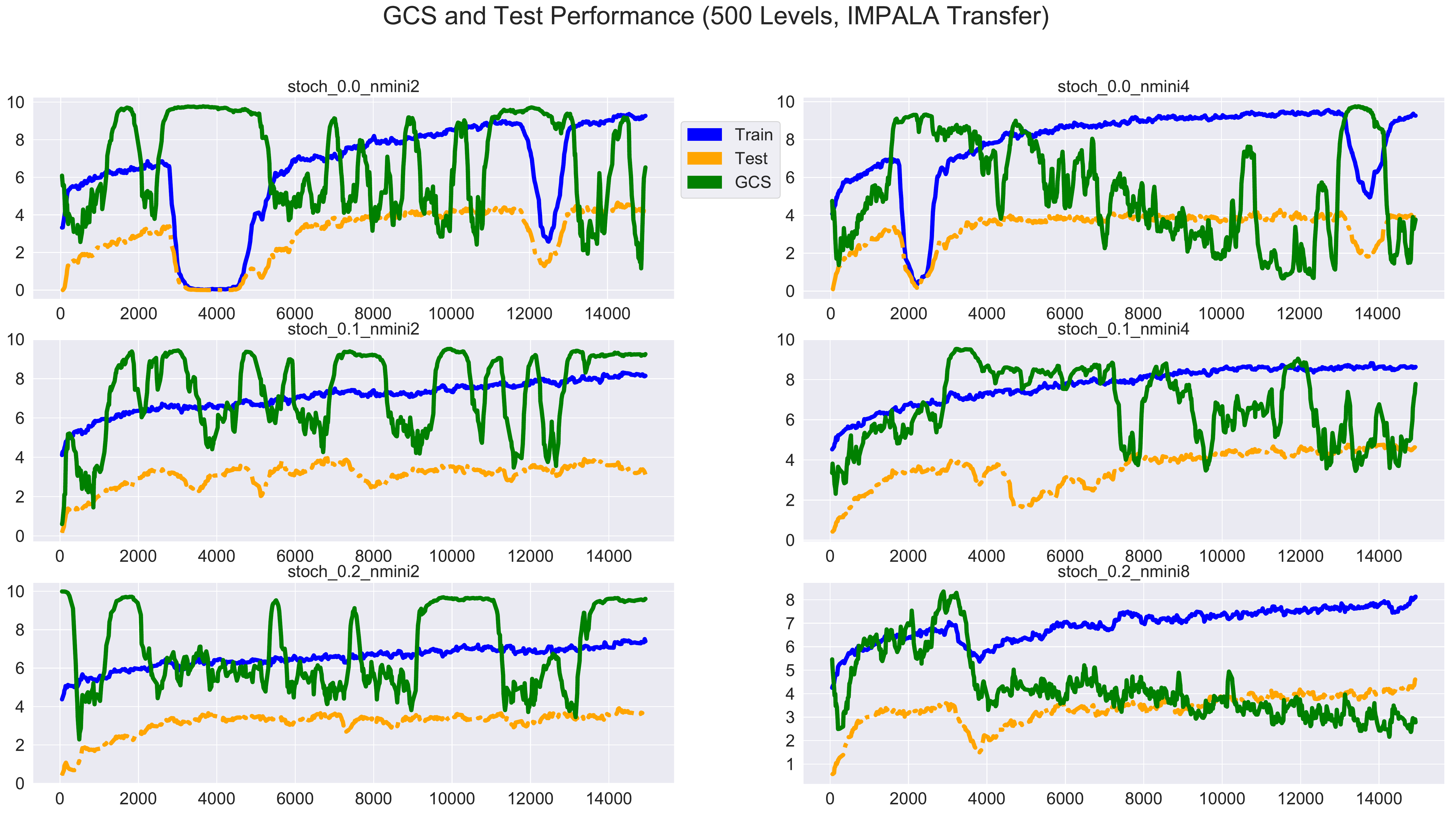}
    \label{fig:gcs_unstable} 
\end{figure}
\newpage

\subsubsection{When is GCS Inaccurate?}
We present an extreme case in which GCS correlation with test performance is diminished: higher complexity on the reward function. If using a synthetic reward (i.e. k-layer convolution networks on the observation, and then average-pooling on the output), we find a monotonic decrease in GCS, but testing performance varied if using the same number of environments for each of the training sets:
\begin{figure}[H]
  \caption{Varying $k$ for Synthetic Reward.} 
  \centering
    \includegraphics[scale = 0.15]{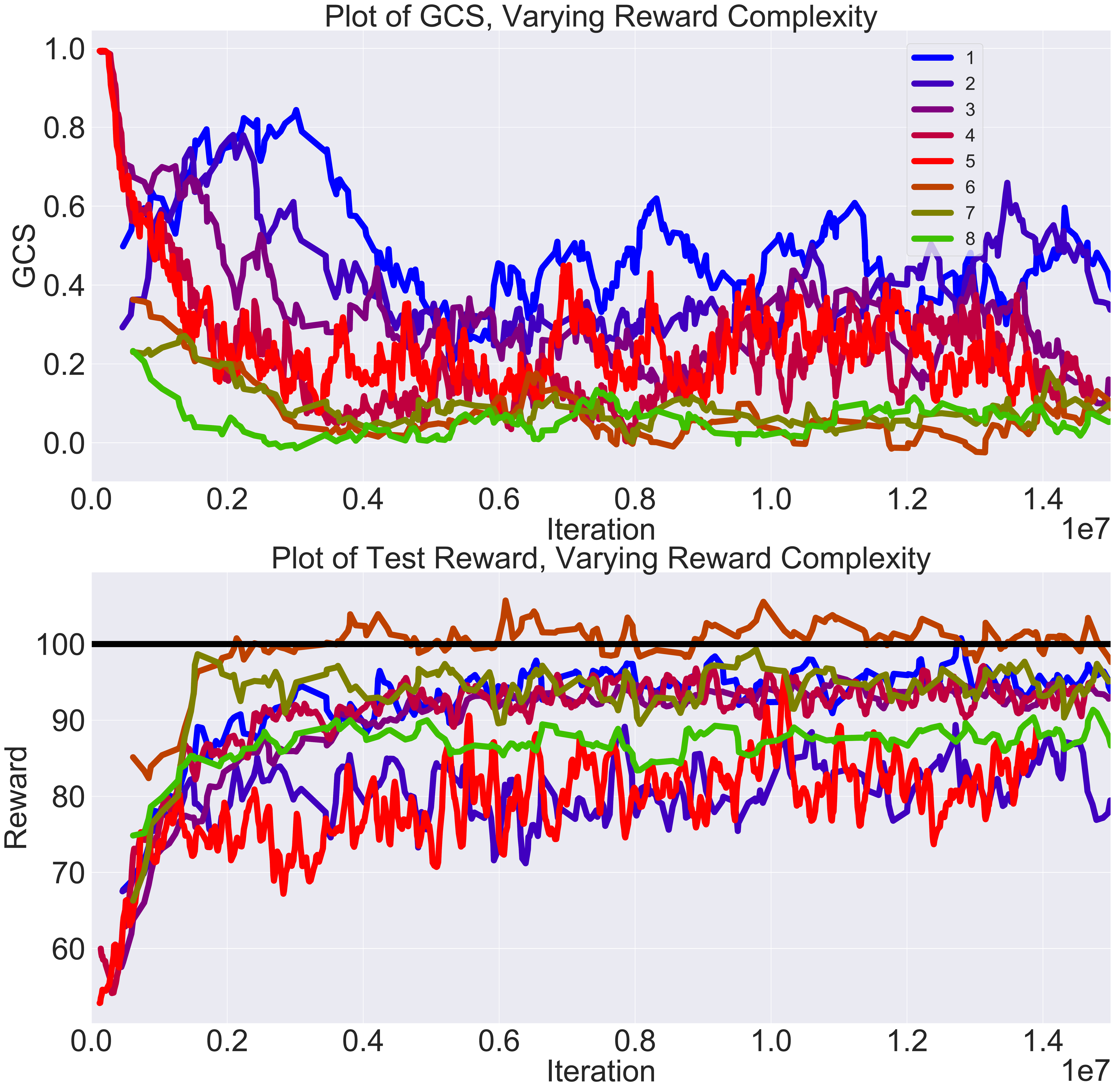}
\end{figure}
The layer number $k$ induces more complexity on the reward function.

\newpage 
\subsubsection{Gradient Norm}
\begin{figure}[H]
  \caption{Gradient Norm vs Training Reward for Various Stochasticities. Note: x-axis uses earliest time for which the reward was reached, then takes the gradient norm and GCS at that moment in time. }
  \centering
    \includegraphics[scale = 0.2]{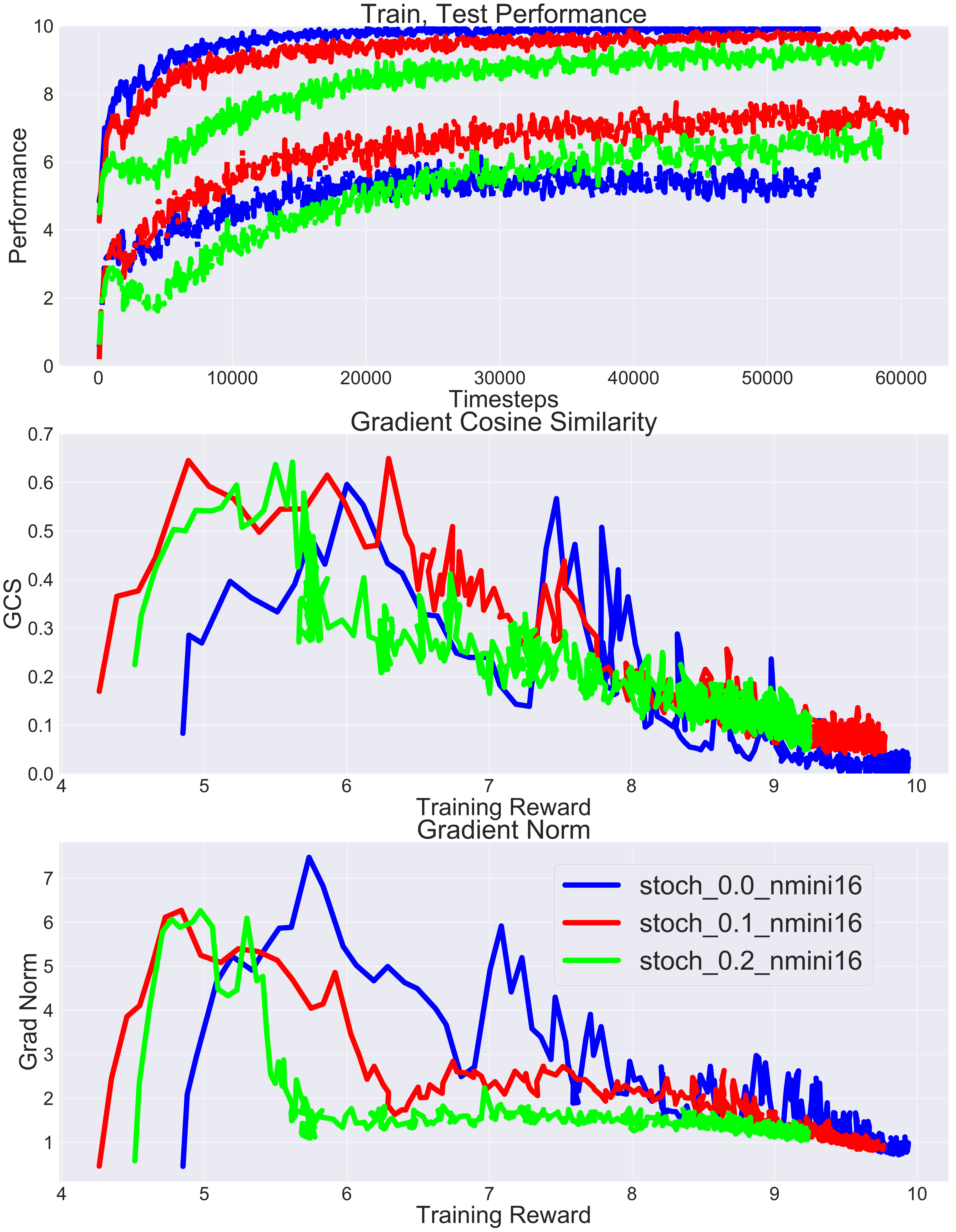}
    \label{fig:stoch_gradnorm}
\end{figure}

We examine the contributions to the GCS, with
(\ref{fig:stoch_gradnorm}) further showing that stochasticity reduces the gradient norm significantly, even while the policy has not yet converged.
\newpage 
\subsubsection{$\gamma$, Continued}
\begin{figure}[H]
  \caption{Varying $\gamma$ for nonstochastic and stochastic CoinRun.} 
  \centering
    \includegraphics[scale = 0.23]{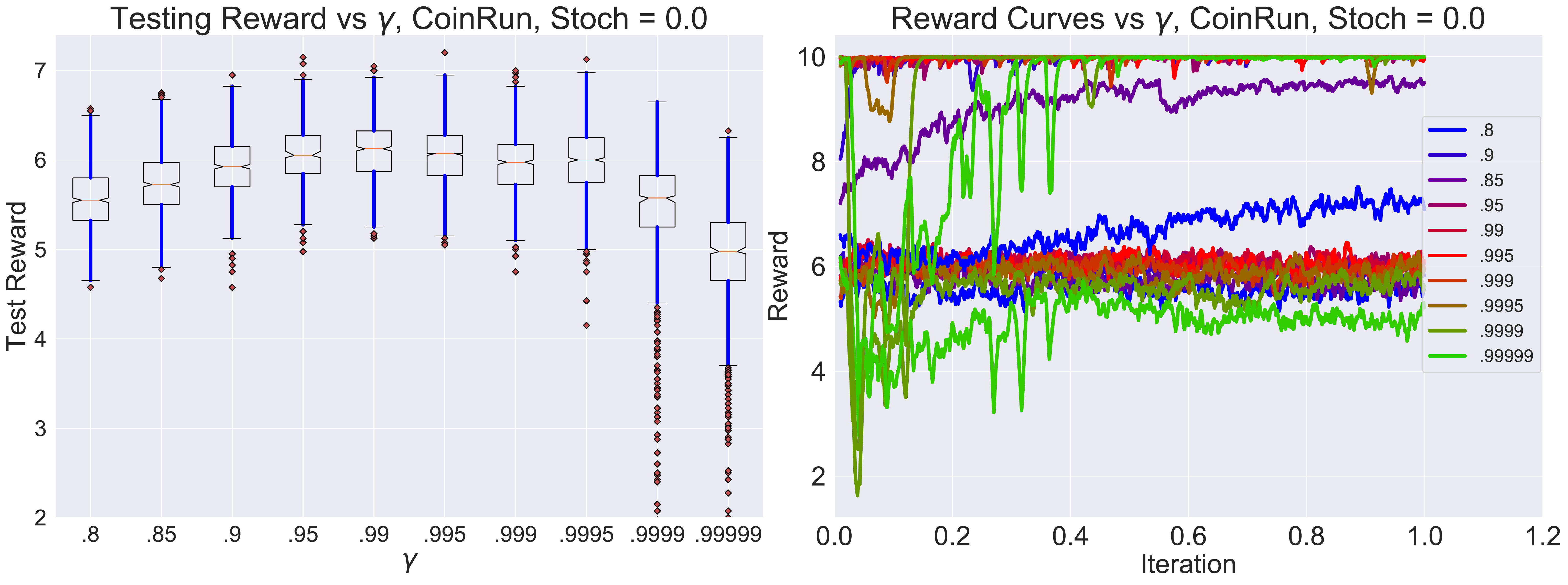}
    \includegraphics[scale = 0.23]{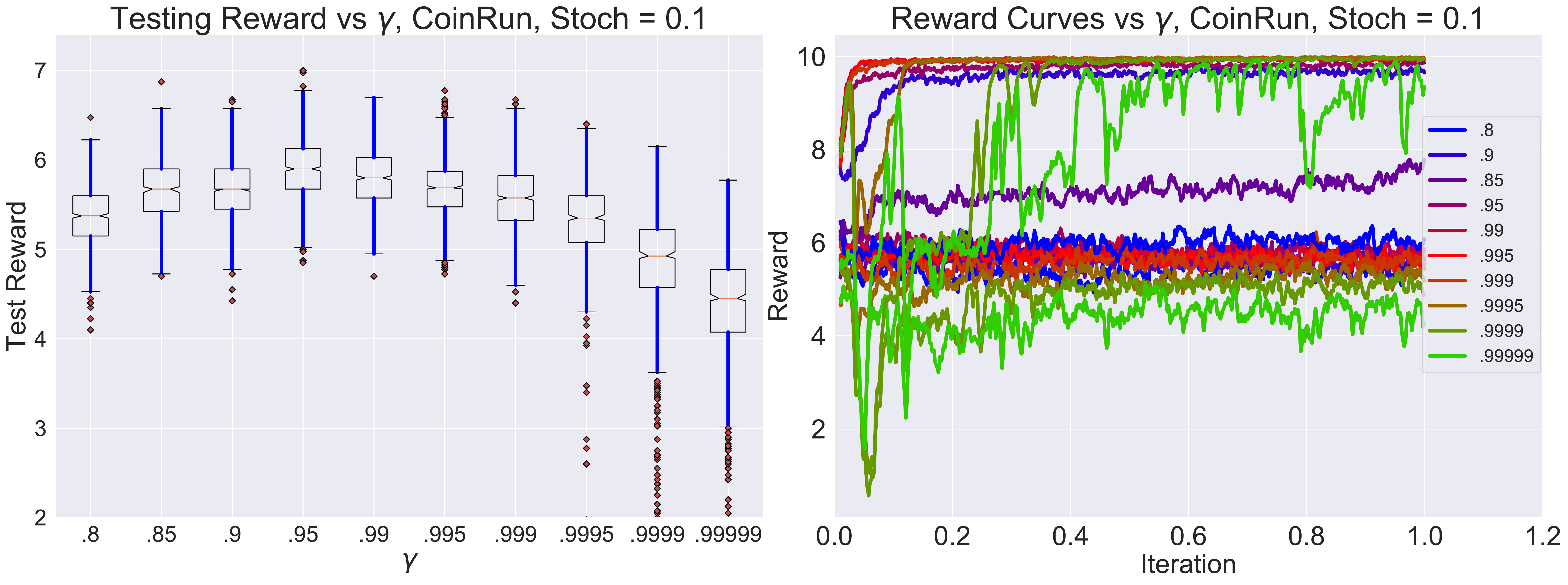}
    \includegraphics[scale = 0.23]{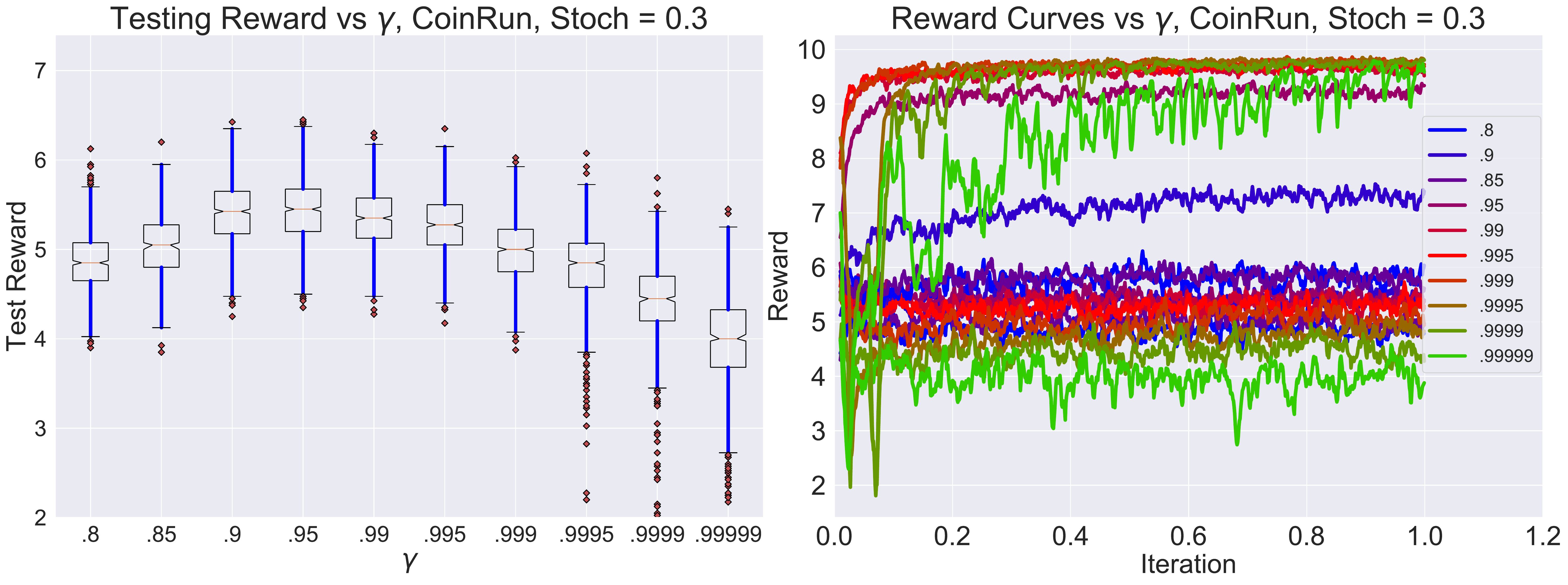}
    \label{fig:gammastoch_full}
\end{figure}

In order to check on the role between $\gamma$ and stochasticity at a larger scale, we apply forced action stochasticity in both \textit{training and testing} settings but vary the $\gamma$, so that there is only one difference between training and testing. (\ref{fig:gammastoch_full}) shows entropy noticeably affecting the range of $\gamma$, where higher stochasticity sharpens the range of allowable $\gamma$'s for high test performance.

\newpage 
\subsubsection{CoinRun-Mazes}
Applying the same stochasticities to the RandomMazes environment which is a test of exploration strategy in a Maze, and we find that stochasticity actually improves the larger $\gamma$ ranges for test performance - exploration/RNN tasks have more complexity due to the temporal component, consistent with this result.

\newpage
\section{Hyperparameters and Exact Setups}
\subsection{RNN-MDP}

\begin{center}
    \begin{tabular}{ | l | l |}
    \hline
    RNN MDP HyperParameters & Values \\ \hline \hline
    RNN Cell  & GRU-512  \\ \hline 
    Horizon & 256 \\ \hline
    State Noise & Gaussian Vector \\ \hline
    Reward Output  & 2-layer MLP ReLU    \\ \hline 
    Policy Function & 2 to 4 layer MLP ReLU \\ \hline
    Initial State Sampling  & Gaussian Vector   \\ \hline
    Initializers & Orthogonal Initializers for all \\ \hline  
    Length of reward gradient & 40 \\ \hline
    Optimizer & SGD or Full-Batch GD \\ \hline

    \end{tabular}
\end{center}

\subsection{PPO Parameters} \label{hyper}
See \cite{coinrun} for the default parameters used for CoinRun.

\end{document}